\RequirePackage{silence}
\PassOptionsToPackage{table}{xcolor}

\documentclass{article} 
\usepackage[final]{colm2026_conference}
\usepackage{graphicx}
\usepackage{microtype}
\usepackage{hyperref}
\usepackage{bookmark}
\usepackage{url}
\usepackage{amsmath}
\usepackage{soul}

\usepackage{spverbatim}
\usepackage{booktabs}
\usepackage{xcolor}
\usepackage{array}
\usepackage{enumitem}
\usepackage{arydshln}
\usepackage{multicol}
\usepackage{multirow}
\usepackage{tabularx}
\usepackage{makecell}
\definecolor{success-green}{RGB}{235, 250, 235}
\definecolor{failure-red}{RGB}{255, 240, 240}
\definecolor{header-gray}{RGB}{240, 240, 240}
\definecolor{proto-blue}{RGB}{245, 248, 255}

\usepackage{siunitx}
\usepackage{lineno}

\definecolor{darkblue}{rgb}{0, 0, 0.5}
\hypersetup{colorlinks=true, citecolor=darkblue, linkcolor=darkblue, urlcolor=darkblue}

\DeclareRobustCommand{\method}{RELISH}

\title{\new{{\method}: LLM \underline{RE}gression with a \underline{L}atent \underline{I}terative \underline{S}tate \underline{H}ead}}

\author{Yiheng Su \\
School of Information \\
The University of Texas at Austin \\
\texttt{sam.su@utexas.edu} \\
\And
Matthew Lease \\
School of Information \\
The University of Texas at Austin \\
\texttt{ml@utexas.edu} \\
}

\definecolor{revision-purple}{RGB}{126, 34, 206}
\newif\ifshowchanges
\showchangesfalse
\DeclareRobustCommand{\new}[1]{%
  \ifshowchanges%
    \textcolor{revision-purple}{#1}%
  \else%
    #1%
  \fi%
}

\begin{document}

\ifcolmsubmission
\linenumbers
\fi

\maketitle

\begin{abstract}
We present \textbf{\method} (\textbf{RE}gression with a \textbf{L}atent \textbf{I}terative \textbf{S}tate \textbf{H}ead), a novel, lightweight architecture designed for text regression with large language models. Rather than decoding numeric targets as text or aggregating multiple generated outputs, {\method} predicts scalar values directly from frozen LLM representations by iteratively refining a learned latent state through cross-attention over token-level representations, and then mapping the final state to a point estimate with a linear regressor. Across \new{six} datasets, four LLM backbones, and two LLM training regimes, {\method} consistently outperforms prior baselines from all three major LLM regression families, including autoregressive decoding, regression-aware inference, and existing predictive head methods. Despite these gains, {\method} remains highly parameter-efficient, 
requiring only $\sim$3.4–3.7M trainable
parameters across frozen LLM backbones (only 0.01–0.04\% additional overhead), far less than LoRA-based alternatives that grow with model size
(0.26–0.42\%). \new{Our code is available at \url{https://github.com/SamSoup/RELISH}}.

%
\end{abstract}

\section{Introduction}


In the era of large language models (LLMs), most natural language processing (NLP) tasks are unified under the \textit{text-to-text} paradigm, in which models consume text inputs and produce text outputs \citep{LiuT52020, anthropic2025system, singh2025openai, yang2025qwen3, google2026gemini31}. Theoretically, this paradigm is appealing because the same pre-trained model, linguistic interface, and decoding procedure can be applied across diverse applications that traditionally required different architectures. Empirically, this paradigm has been particularly successful for intrinsically \textit{generative} tasks such as question answering, summarization, translation, and code generation \citep{Clark2018-ri, Hendrycks2020-cb, Chen2021-nq, srivastava2023beyond, NEURIPS2024_ad236edc, kazemi-etal-2025-big, Li2025-zf, Center-for-AI-Safety2026-sh}. Here, the need to produce coherent, contextually appropriate text aligns naturally with the language modeling objectives used to pre-train LLMs, such as autoregressive next-token prediction \citep{radford2019language, Brown2020-ni}. 


However, the \textit{text-to-text} paradigm is less suited to \textit{predictive} tasks such as classification and regression \citep{Lukasik2025-ak}. 
Unlike generative tasks whose outputs are naturally text, predictive tasks seek discrete labels or continuous scalars.
In regression, this mismatch makes text generation suboptimal because predictive quality depends on numerical error, whereas language modeling penalizes token mismatches. 
For example, if the target is 1.0, a regression objective like MSE prefers 0.9 to 0.1, whereas a language modeling objective like cross-entropy treats them as discrete tokens and remains agnostic to numerical proximity.
Recent work has explored how to adapt LLMs to regression. Three major families have emerged (see Table~\ref{tab:regression-taxonomy}): \textit{autoregressive decoding} \citep{vacareanu2024from, Song2024-kc, Song2025-dx, Akhauri2025-rf}, \textit{regression-aware inference} \citep{Lukasik2024-yt, Lukasik2025-ak, Chiang2025-ax}, and \textit{predictive heads} \citep{xin2021berxit, Wang2022-ko, Zhuang2023-om, Fernandes2023-ty, Zhang2024-tz, tang2025understanding, nguyen2024predicting}. 
Autoregressive methods simply prompt the LLM. 
Regression-aware methods 
complement next-token generation with regression-aware decision rules that explicitly minimize numerical error. Predictive heads avoid text generation entirely by finetuning a predictive head (e.g., a linear layer) attached to the LLM to directly output numerical estimates. 




Autoregressive decoding is simplest and most faithful to how LLMs are pre-trained \citep{vacareanu2024from}. Consequently, it is least aligned with the numerical nature of regression targets, since decoding is based on token probabilities rather than numerical proximity. Regression-aware inference seeks to remedy this mismatch by forming predictions from the LLM’s output distribution using Bayes-optimal decision rules to directly minimize the loss metric, such as the posterior mean under mean squared error \citep{Lukasik2024-yt, Lukasik2025-ak}. However, approximating this posterior often requires sampling multiple outputs or enumerating candidate targets, thereby increasing computation at inference time. Moreover, this approach only partially resolves the mismatch, since the posterior is still defined over textualized numbers (e.g., ``3.5'') rather than directly over the continuous output space.

Predictive head methods instead bypass text generation at inference time by mapping LLM representations directly to numeric predictions through a task-specific regressor. This enables direct optimization in the continuous output space, yielding efficient single-pass inference. However, existing approaches \citep{Zhuang2023-om, Zhang2024-tz, nguyen2024predicting, tang2025understanding} are limited, typically relying on basic pooling heuristics (e.g., mean pooling) over frozen hidden states. As a result, they may fail to fully extract regression-relevant information from LLM representations and thus may underperform regression-aware inference. \textit{Can we match the predictive performance of regression-aware approaches without sacrificing the efficiency of predictive heads?}


In this work, we propose \textbf{RE}gression with a \textbf{L}atent \textbf{I}terative \textbf{S}tate \textbf{H}ead (\textbf{{\method}}), a novel, lightweight yet expressive predictive head architecture for LLM-based regression. Rather than decoding numeric targets as text or aggregating across several generated outputs, {\method} predicts scalar values directly from frozen LLM representations through an iterative refinement mechanism. Concretely, {\method} repeatedly updates a learned latent state via cross-attention over the LLM’s token representations, and then maps the final state to a scalar prediction with a linear regressor. Compared with existing predictive heads based on simple pooling heuristics, {\method} provides a more robust summarization mechanism for extracting regression-relevant information without sacrificing inference efficiency. 


\new{
We evaluate {\method} on six datasets spanning semantic textual similarity \citep{marelli-etal-2014-sick, muennighoff-etal-2023-mteb}, machine translation quality estimation \citep{specia-etal-2020-findings-wmt}, and code quality estimation \citep{Song2026StatLLM}. 
While the first two are canonical regression tasks, evaluating LLM-generated code has gained prominence given the rise of autonomous agents.
In AI4Science \citep{zheng-etal-2025-automation}, recent benchmarks valuably assess if generated code produces correct scientific outputs \citep{ScienceAgentBench2025, astrovisbench2025}, but estimating code quality at inference time, necessary for calibrated trust, remains underexplored.
}

\new{Empirically, {\method} consistently outperforms state-of-the-art regression baselines across the six datasets, four LLM backbones (8--32B), and both frozen and LoRA \citep{hu2022lora} LLM training regimes.}
Relative to the second-best method, RAFT \citep{Lukasik2025-ak}, {\method} improves Pearson correlation 
\new{by 10.2\%,}
Spearman correlation 
\new{by 9.3\%,} and range-normalized RMSE 
\new{by 19.7\%}
(Table~\ref{tab:relish_overall_main_with_statllm}). 
{\method} is also highly parameter-efficient, requiring only $\sim$3.4–3.7M trainable
parameters across frozen LLM backbones (only 0.01–0.04\% more overhead), far less than LoRA-based alternatives (Table~\ref{tab:lightweight}) that grow with model size
(0.26–0.42\%).
%
\textbf{Overall, our key contributions are}: 
\begin{enumerate}[topsep=0pt, itemsep=0pt, parsep=2pt]
    \item A novel predictive head architecture designed for text regression ($\S$~\ref{sec:method}). 

    \item A unified conceptual ($\S$~\ref{sec:related_inference}) and empirical ($\S$~\ref{sec:results}) comparison of autoregressive decoding, regression-aware inference, and predictive head methods across \new{six} datasets, four LLM backbones, and both frozen and LoRA fine-tuned regimes, showing that {\method} achieves the strongest overall performance with a substantially smaller trainable footprint than LoRA.

\end{enumerate}
\vspace{-1em}


\begin{table*}[t]
\centering
\small
\renewcommand{\arraystretch}{1.4} 
\begin{tabularx}{\textwidth}{@{} >{\raggedright\arraybackslash\bfseries}p{0.18\textwidth} >{\hsize=1.13\hsize}X >{\hsize=0.87\hsize}X @{}}
\toprule
\textbf{Inference family} & \textbf{How to make predictions} & \textbf{Suitable training setups} \\ \midrule

Autoregressive \newline decoding \newline \citep{vacareanu2024from} & 
\textit{Parse from string:}\vspace{4pt} \newline
$\displaystyle \hat{y}_{\textsc{ard}}(x)=\text{float}(\arg\max_{y\in\mathcal{Y}} p(\text{str}(y)\mid x))$ & 
\textbf{PEFT/SFT:} Minimize token-level CE on $\text{str}(y^*)$.
\vspace{4pt} \newline
$\displaystyle \mathcal{L}_{\textsc{ce}} = -\sum_{t=1}^{|str(y^*)|}\log p(y_t \mid x, y_{<t})$ 
\\ \addlinespace

Regression-aware inference \newline (RAIL) \newline \citep{Lukasik2024-yt} & 
\textit{Bayes-optimal from $p(\cdot\mid x)$:}\vspace{4pt} \newline
$\displaystyle \hat{y}_{\textsc{rail}}(x) = \arg\min_{v\in\mathbb{R}} \mathbb{E}_{y\sim p}[\ell(\text{float}(y),v)]$ \vspace{6pt} \newline
When $\ell = \mathcal{L}_{\textsc{mse}}$, same as posterior mean:\vspace{4pt} \newline
$\displaystyle \hat{y}_{\textsc{rail}}(x) \approx \sum_{y\in\mathcal{Y}_{\text{grid}}} y \cdot p(\text{str}(y)\mid x)$ & 
\textbf{PEFT/SFT:} \vspace{2pt} \newline
(i) Minimize $\mathcal{L}_{\textsc{ce}}$. \vspace{4pt} \newline
(ii) \textit{RAFT} \citep{Lukasik2025-ak}. \vspace{4pt} \newline
$\displaystyle \mathcal{L}_{\textsc{raft}} = \Big(y^* - \mathbb{E}_{y\sim p}[\text{float}(y)]\Big)^2$ \\ \addlinespace

Predictive Head \newline \citep{tang2025understanding} & 
\textit{Extract representations, then regress:}\vspace{4pt} \newline
$\displaystyle \hat{y}_{\textsc{head}}(x)=f_{\textsc{reg}}\big(\phi\big(H(x)\big)\big)$ \vspace{6pt} & 
\textbf{Head-only:} train $f_{\textsc{reg}}$ with $\mathcal{L}_{\textsc{mse}}$. \newline
\textbf{PEFT/SFT + head:} fine-tune backbone jointly with $f_{\textsc{reg}}$. 
\\ \addlinespace

\method \newline (ours) & 
\textit{Iteratively refine a latent state, then regress:}\vspace{2pt} \newline
$\displaystyle r^{(i)}=\mathrm{Refine}\!\left(r^{(i-1)},H(x)\right), \ i=1,\dots,L$ \vspace{6pt} \newline
$\displaystyle \hat{y}_{\textsc{\method}}(x)=w^\top r^{(L)}+b$ & 
\textbf{Head-only:} train $r^{(0)}$, $w$, $b$, and refinement modules with $\mathcal{L}_{\textsc{MSE}}$. \vspace{2pt} \newline
\textbf{PEFT/SFT + head:} optionally fine-tune backbone jointly with RELISH modules. 
\\

\bottomrule
\end{tabularx}
\caption{\textbf{Taxonomy of methods leveraging LLMs for regression.}
We characterize LLM-based regression along two axes: (a) how the model produces a point estimate (\textit{inference family}) and (b) which parameters are trained and how (\textit{training setup}). Autoregressive decoding generates $\text{str}(y)$ and parses it. RAIL computes a Bayes-optimal estimate under $p(\cdot\mid x)$ for loss $\ell$ (e.g., the posterior mean for $\mathcal{L}_{\textsc{mse}}$), where the posterior is estimated via sampling or candidate enumeration over a grid $\mathcal{Y}_{\text{grid}}$. Predictive heads regress from pooled hidden states, whereas \method{} replaces static pooling with iterative latent-state summarization over token-level representations. Typical objectives are $\mathcal{L}_{\textsc{ce}}$ for autoregressive decoding, $\mathcal{L}_{\textsc{ce}}$ or $\mathcal{L}_{\textsc{raft}}$ for RAIL, and $\mathcal{L}_{\textsc{mse}}$ for predictive-head methods.}
\label{tab:regression-taxonomy}
\end{table*}

\section{Background on LLM-based Regression}
\label{sec:background}
\vspace{-0.5em}

In this section, we begin with problem formulation ($\S$~\ref{sec:problem_setup}), then review existing methods for 
inference ($\S$~\ref{sec:related_inference}) and training ($\S$~\ref{sec:related:training}) (see Table~\ref{tab:regression-taxonomy}). Last, we differentiate {\method} in $\S$~\ref{sec:relish_diff}. 

\subsection{Problem Formulation}
\label{sec:problem_setup}

Following \citet{Lukasik2025-ak}, we study \textbf{natural language regression}. Given a dataset $\mathcal{D}=\{(x_i,y_i^*)\}_{i=1}^N$, where each $x_i$ is a textual input and each $y_i^* \in \mathbb{R}$ is a continuous target, the goal is to learn a predictor $\hat y(x) \in \mathbb{R}$ that minimizes some regression loss $\mathcal{L}(y^*,\hat y)$.



LLMs offer a flexible foundation for natural language regression. They can encode rich textual inputs without manual feature engineering, transfer broad semantic knowledge from pre-training, and support a common interface across diverse tasks \citep{Song2024-kc, Akhauri2025-rf}. However, \textit{there is no consensus on how to best adapt LLMs for regression}. 

We categorize prior work by two axes:
(a) \textbf{how the model produces a numeric prediction}, and
(b) \textbf{whether (and how) the model is trained}.
For (a), three inference families emerge: autoregressive decoding, regression-aware inference, and predictive head methods. For (b), four training regimes emerge: no training (prompt-only), parameter-efficient fine-tuning (PEFT), full supervised fine-tuning (SFT), and training only the predictive heads while freezing the LLM backbone. \textbf{Table~\ref{tab:regression-taxonomy}} summarizes this taxonomy, and \textbf{Table~\ref{tab:operational_tradeoffs}} compares the operational costs of standard LLM regression approaches.

Inference methods and training regimes are largely orthogonal. For instance, the same inference procedure (e.g., predictive heads) can be applied to either a frozen or fine-tuned LLM, while the same training regime (e.g., PEFT) can support different inference procedures. However, some pairings are more natural than others, since certain training objectives (e.g., regression-aware objectives) better support particular inference procedures.

\begin{table}[t]
\centering
\small
\setlength{\tabcolsep}{5pt} 
\renewcommand{\arraystretch}{1.2} 
\begin{tabular}{@{} l l c c @{}}
\toprule
\textbf{Method} & \textbf{Family} & \textbf{Training} & \textbf{Inference} \\
\midrule
Zero / Many Shot \citep{vacareanu2024from} & Autoregressive & None & \new{Medium--High} \\
Linear / MLP \citep{tang2025understanding} & Predictive head & Low & \new{Low} \\
RAIL \citep{Lukasik2024-yt} & Regression-aware & None & \new{High} \\
RAFT \citep{Lukasik2025-ak} & Regression-aware & High & \new{High} \\
\textbf{RELISH (ours)} & \textbf{Predictive head} & \textbf{Low} & \new{\textbf{Low}} \\
\bottomrule
\end{tabular}
\caption{\textbf{Qualitative operational costs of common LLM regression approaches.} 
\new{
Training costs reflect optimization overhead. Autoregressive methods and RAIL do not update parameters (None), predictive heads and RELISH freeze the LLM backbone and only train auxiliary modules (Low), while RAFT fine-tunes the LLM backbone (High). Inference costs correspond to runtime latency. Predictive heads and RELISH are lightweight (Low), whereas autoregressive methods sequentially decode tokens, so latency scales with the number of in-context demonstrations (Medium--High). RAIL and RAFT approximate model posterior via multiple samples or scoring grid candidates (High).
}
}
\label{tab:operational_tradeoffs}
\end{table}

\subsection{Inference Paradigms}
\label{sec:related_inference}

\paragraph{Autoregressive decoding (ARD).}
Autoregressive decoding \citep{vacareanu2024from, Song2024-kc, Song2025-dx, Akhauri2025-rf} treats scalar targets as text. Given an input $x$, a decoder LLM produces a point prediction by sequentially generating a numeric string from vocabulary tokens $\mathcal{Y}$ and then parsing it into a real value:
\begin{equation*}
\hat{y}_{\text{ARD}}(x) = \text{float}(\arg\max_{y \in \mathcal{Y}} p(\text{str}(y) \mid x))
\end{equation*} \vspace{-1em}

In practice, the most common approach is direct prompting, either zero-shot or with many-shot demonstrations. Thus, ARD requires no task-specific training, \new{although inference overhead scales with the number of in-context demonstrations} (Table~\ref{tab:operational_tradeoffs}). 
\new{Additionally}, \new{ARD is sensitive to prompt formatting and insensitive to numerical proximity among candidate predictions \citep{Lukasik2025-ak}.}



\paragraph{Regression-aware inference (RAIL).}
RAIL \citep{Lukasik2024-yt} addresses the representational mismatch by \new{augmenting} autoregressive decoding with a loss-aware decision rule over the LLM’s predictive distribution. Formally, given an input $x$, RAIL predicts:
\begin{equation*}
\hat{y}_{\textsc{RAIL}}(x) =
\arg\min_{v\in\mathbb{R}}
\mathbb{E}_{y\sim p(\cdot\mid x)}
\left[\ell\left(\mathrm{float}(y), v\right)\right].
\end{equation*}
This is equivalent to the Bayes-optimal decision rule for the target loss $\ell$ \citep{kumar-byrne-2004-minimum}. For mean squared error, the optimal prediction is the mean of the posterior. In practice, this posterior is approximated either by sampling multiple completions or, when feasible (e.g., ordinal regression), by enumerating a grid of candidate values.

As summarized in Table~\ref{tab:operational_tradeoffs}, RAIL thus trades higher inference cost \new{to better align textual predictions with numerical proximity}.
\citet{Lukasik2024-yt} show that RAIL improves over greedy and self-consistency decoding \citep{wang2023selfconsistency} on semantic text similarity and sentiment analysis tasks. 
Subsequent work further strengthens RAIL through regression-aware fine-tuning \citep{Lukasik2025-ak}, which we discuss in Section~\ref{sec:related:training}.

\paragraph{Predictive heads.}
Predictive heads \citep{Fernandes2023-ty, Zhuang2023-om, nguyen2024predicting, tang2025understanding} avoid text generation by predicting continuous scalars directly from internal LLM representations. Given an input $x$, let $H(x)\in\mathbb{R}^{S\times d}$ denote the token-level hidden states of the LLM, where $S$ is the sequence length and $d$ is the LLM's hidden dimension. Predictive head methods first summarize token representations into a fixed-size vector, and then apply a downstream regressor:
\begin{equation*}
\hat{y}_{\textsc{head}}(x)=f_{\textsc{reg}}(\phi(H(x))).
\end{equation*}
Here, $\phi$ typically \new{represents a static compression heuristic}, such as mean-pooling across token sequences \citep{li2020sentence} or indexing the representation of a specific token, such as \texttt{[CLS]} \citep{reimers-gurevych-2019-sentence}. The regressor $f_{\textsc{reg}}$ varies by implementation. \textit{Vector-based} heads employ an auxiliary module, such as a dense linear layer \citep{Zhuang2023-om} or a 2-layer MLP \citep{tang2025understanding}. \textit{Logit-based} heads repurpose the LLM's language modeling head by extracting the logit of a designated vocabulary token \citep{Fernandes2023-ty, Zhuang2023-om} to emit as predictions. 

\subsection{Training Regimes}
\label{sec:related:training}

Complementary to the inference mechanism, LLM regression methods also differ in \emph{which parameters are updated} and \emph{which loss objective is used}. As summarized in Table~\ref{tab:regression-taxonomy}, these choices are largely separable from the inference family, though some objectives are more naturally paired with certain prediction rules.

\paragraph{No training.}
The simplest regime uses the LLM as-is, so all task adaptation occurs through the prompt rather than gradient updates. This regime is most natural for autoregressive decoding, but can also be combined with regression-aware inference by applying a regression-aware decision rule to the output distribution of a frozen model (e.g., RAIL).

\paragraph{Head-only training.}
\new{Predictive heads typically freeze the LLM backbone and only train auxiliary head parameters.} Vector-based heads update the regressor $f_{\textsc{reg}}$, while logit-based heads update the linear matrix $W \in \mathbb{R}^{d \times |\mathcal{Y}|}$ that projects model hidden states to vocabulary tokens. \new{Since predictive heads emit continuous scalars, they can directly optimize numerical error using classic regression loss functions, such as mean squared error ($\mathcal{L}_{\textsc{mse}}$).}


\paragraph{LLM backbone adaptation: PEFT and SFT.}
The LLM backbone can be adapted via parameter-efficient fine-tuning (e.g., LoRA \citep{hu2022lora}) or full supervised fine-tuning. For autoregressive decoding, the model is typically optimized using token-level cross-entropy to reproduce the textualized target string exactly:
\begin{equation*}
\mathcal{L}_{\textsc{ce}} = -\sum_{t=1}^{|str(y^*)|}\log p(y_t \mid x, y_{<t}).
\end{equation*}
While $\mathcal{L}_{\textsc{ce}}$ does not directly minimize numerical error, large-scale training can lift generation to match or exceed scalar alternatives \citep{Song2024-kc, Song2025-dx, Akhauri2025-rf, akhauri2026regression}. However, these methods rely on massive proprietary datasets (e.g., telemetry logs) to learn numerical continuity and are impractical for most applications. 



\paragraph{Regression-aware fine-tuning.}
For regression-aware inference, one may still fine-tune the LLM with $\mathcal{L}_{\textsc{ce}}$, but a more natural objective is to directly optimize the same posterior statistic used at runtime. RAFT \citep{Lukasik2025-ak} operationalizes this by minimizing:
\begin{equation*}
\mathcal{L}_{\textsc{raft}} = \Big(y^* - \mathbb{E}_{y\sim p}[\mathrm{float}(y)]\Big)^2,
\end{equation*}
thereby aligning training with the Bayes-optimal estimator used by RAIL. This is computationally more expensive, since the posterior must be approximated during training as well, and inference still requires multiple samples or candidate enumeration (see Table~\ref{tab:operational_tradeoffs}).

\subsection[How is RELISH different?]{How is \method{} different?}
\label{sec:relish_diff}



\new{
Similar to existing predictive heads, {\method} also compresses token representations into a fixed-size vector for regression, preserving high training and inference efficiency. 
However, prior predictive heads are limited by an \textit{information extraction bottleneck}. Specifically, vector-based methods collapse token representations using task-agnostic heuristics such as mean pooling, which can wash out localized or distributed signals before the downstream regressor sees them. 
Logit-based heads avoid explicit pooling, but bind the prediction to a designated vocabulary logit at a specific output position. 
This architecture forces the model to encode all regression-relevant signals into that output position's representation. Consequently, any localized or distributed cues omitted from this representation cannot be recovered at prediction time. 
}

In contrast, {\method} \new{mitigates this information extraction bottleneck} with an iterative refinement mechanism, \new{in which the predictive head repeatedly updates a learned summary of token representations for regression.} \new{By repeatedly attending over all token representations, this learned summary can retain localized cues while accumulating distributed signals spread across multiple tokens.}
Concretely, {\method} maintains a learned latent state $r^{(0)}$ and repeatedly refines $r$ through cross-attention over the token representations $H(x)$:
\begin{equation*}
r^{(i)}=\mathrm{Refine}\!\left(r^{(i-1)}, H(x)\right), \qquad i=1,\dots,L,
\end{equation*}
The final prediction is then emitted by a linear regressor:
\begin{equation*}
\hat{y}_{\textsc{\method}}(x)=w^\top r^{(L)}+b.
\end{equation*}
By leveraging trainable latent queries rather than static pooling or indexing, {\method} is related in spirit to DeepMind's Perceiver IO \citep{jaegle2022perceiver} and BLIP-2 architectures \citep{blip2023}. 
Unlike prior latent-query architectures designed for general encoding or multimodal bridging, {\method} is specifically engineered for regression by iteratively distilling token-level hidden states into a single latent state optimized for scalar prediction.
As summarized in Table~\ref{tab:operational_tradeoffs}, our design preserves \new{the training and inference efficiency} of predictive heads while providing a more robust summarization mechanism.

\new{
Beyond efficiency, {\method} offers conceptual advantages over autoregressive and regression-aware methods by directly predicting scalars from LLM representations. 
Autoregressive methods serialize continuous targets into discrete tokens, but token likelihoods do not naturally encode numerical proximity. 
Regression-aware methods attempt to mitigate this limitation by estimating Bayes-optimal statistics from the LLM's induced output distribution. 
Collectively, these methods essentially attempt to calibrate probability mass over serialized numeric candidates. 
Consequently, predictions may be sensitive to the \textit{verbalization gap} \citep{yuchi-du-eisner-2026}, tokenization artifacts, candidate-set design, and learned semantic associations from pre-training. 
$\S$~\ref{sec:app_expanded_related} provides additional details on how {\method} differentiates from existing LLM-based regression methods.
}


\section{Method}
\label{sec:method}

\subsection{Setup}
\new{
Given an input sequence $x$ of length $S$, we extract token-level representations from the penultimate layer (see $\S$~\ref{sec:app_penultimate_layer} for justification) of a frozen decoder-only LLM:
}
\[
H(x)=\mathrm{LLM}_{\mathrm{frozen}}(x)\in\mathbb{R}^{S\times d}.
\]
Following \citet{nguyen2024predicting}, \citet{Song2024-kc}, and \citet{tang2025understanding}, we apply y-normalization to standardize \new{ground-truth labels} ($y^*$) using training-split statistics $(\mu_y,\sigma_y)$. \new{We train the model to predict} an unbounded standardized target:
\[
z=\frac{y^*-\mu_y}{\sigma_y+\epsilon},
\qquad
\hat y=\hat z(\sigma_y+\epsilon)+\mu_y,
\]
where $\epsilon>0$ is a small constant for numerical stability.

\subsection{Architecture}
{\method} \new{mitigates the information extraction bottleneck of existing predictive heads} with an iterative refinement mechanism over token-level LLM representations. Rather than collapsing the token sequence into a fixed vector in a single step, {\method} maintains a learned latent state that is repeatedly refined via cross-attention over the input tokens before being \new{projected to the final scalar prediction}.

We first project \new{token-level hidden states to} a head dimension $d_h$:
\[
X(x)=H(x)W,\qquad W\in\mathbb{R}^{d\times d_h},
\]
so that $X(x)\in\mathbb{R}^{S\times d_h}$ serves as the token-level memory. This projection is optional if $d_h = d$. See $\S$~\ref{sec:relish_setup_extended} on why we recommend including this step. 

Let $r^{(0)}\in\mathbb{R}^{d_h}$ denote a learned latent state. For $i=1,\dots,L$, {\method} updates this latent state with Transformer-style residual refinement blocks:
\[
\tilde r^{(i)}
=
LN\big(r^{(i-1)}+
\mathrm{MHA}\!\big(r^{(i-1)},X(x),X(x)\big)\big),
\qquad
r^{(i)}
=
LN\big(\tilde r^{(i)}+
\mathrm{FFN}\!\big(\tilde r^{(i)}\big)\big).
\]
Here, LN is a post-layer normalization \citep{Vaswani+2017}, FFN is a feed-forward MLP, and $\mathrm{MHA}(\cdot)$ is a multi-head cross-attention with the latent state as query and token-level representations as keys and values. For each attention head $m$ out of $M$ total heads,
\[
Q_m=r^{(i-1)}W_m^Q,\qquad
K_m=X(x)W_m^K,\qquad
V_m=X(x)W_m^V,
\]
\[
\mathrm{Attn}_m(Q_m,K_m,V_m)=
\mathrm{softmax}\!\left(\frac{Q_mK_m^\top}{\sqrt{d_h/M}}\right)V_m.
\]
After $L$ refinement steps, the final state is mapped to an unbounded standardized prediction:
\[
\hat z=w^\top r^{(L)}+b,
\]
where $w\in\mathbb{R}^{d_h}$ and $b\in\mathbb{R}$ are trainable parameters. 

\new{Although {\method} adopts components from transformers, {\method} is not simply a stack of $L$ additional transformer layers trained on top of the LLM. A standard transformer layer is a \textit{sequence-processing} module. Queries, keys, and values are all derived from the same token sequence $H(x)$, self-attention updates every token representation, and layer propagation aims to output new token sequences $H'(x)$. In contrast, {\method} is a \textit{summarization} module. A {\method} layer uses the learned latent state $r^{(i-1)}$ as the query, which cross-attends to frozen LLM token representations as the keys and values. Each subsequent refinement layer updates only the latent state $r^{(i)}$ and leaves the token representations unchanged.}

\subsection{Training Objective}
We train all {\method} parameters end-to-end while keeping the LLM backbone frozen, i.e., the projection $W$, latent state $r^{(0)}$, refinement blocks, and linear weights $(w,b)$. We optimize the Huber loss \citep{huber1992robust} on standardized targets:
\[
\mathcal{L}_{\mathrm{Huber}}(\hat z,z)=
\begin{cases}
\frac{1}{2}(\hat z-z)^2, & |\hat z-z|\le \delta,\\[2mm]
\delta\big(|\hat z-z|-\frac{1}{2}\delta\big), & \text{otherwise},
\end{cases}
\]
where $\delta>0$ is the Huber threshold.

\section{Evaluation}
\label{sec:main_results}

\subsection{Experimental Setup}
\label{sec:experimental_setup}

\paragraph{LLM Backbones}
We employ four LLM backbones: \textbf{Llama 3.1 8B Instruct} \citep{grattafiori2024llama}, \textbf{Qwen 3 8 and 32B} \citep{yang2025qwen3}, and \textbf{Gemma 3 27B Instruct} \citep{gemma3technicalreport}.

\paragraph{Datasets} We evaluate on \new{three text regression tasks: semantic textual similarity (STS) \citep{enevoldsen2025mmteb}, machine translation quality estimation (WMT) \citep{specia-etal-2020-findings-wmt}, and code quality estimation \citep{Song2026StatLLM}}. For STS, models predict a scalar score representing the degree of semantic equivalence between two sentences. The STS-Benchmark (STS-B) \citep{enevoldsen2025mmteb} contains English sentence pairs from captions and forums on a 0--5 scale. The semantic relatedness subset of Sentences Involving Compositional Knowledge (SICK-R) \citep{marelli-etal-2014-sick} is on a 1--5 scale. \new{Table~\ref{tab:datasets} reports dataset split sizes.}

For WMT, models rate how accurately a machine translation preserves the source language's meaning without human references. The 2020 Multilingual Quality Estimation Task 1 \citep{specia-etal-2020-findings-wmt} provides scores on a 0--100 scale for English--Chinese, Russian--English, and Sinhala--English following \citet{Wang2022-ko}. \new{For code, models assess the correctness and readability of statistical code. The StatLLM dataset \citep{Song2026StatLLM} comprises 207 scientific workflow tasks, for which GPT-3.5 \citep{openai2022chatgpt}, GPT-4 \citep{openai2023gpt4}, and Llama 3.1 70B \citep{grattafiori2024llama} generated SAS code with expert ratings on a 0--50 scale.}

%
%

\paragraph{Metrics} We report Pearson ($r$), Spearman ($\rho$), and range-normalized root mean squared error (NRMSE) in Table~\ref{tab:relish_overall_main_with_statllm}. Because our datasets have different gold label ranges, raw RMSEs are not directly comparable. We therefore report $\mathrm{NRMSE} = \mathrm{RMSE} / (y_{\max} - y_{\min})$, which normalizes prediction error on the same scale across datasets. 
Tables~\ref{tab:stsb_relish_main}, \ref{tab:sickr_sts_relish_main}, \ref{tab:wmt_en_ru_relish_main}, \ref{tab:wmt_en_zh_relish_main}, \ref{tab:wmt_si_en_relish_main}, \new{and~\ref{tab:statllm_code_quality_relish_main}} report raw RMSE scores for each individual dataset. 

\paragraph{Baselines} We compare {\method} against baselines from three LLM-based regression families (Table~\ref{tab:operational_tradeoffs}). For autoregressive decoding, we evaluate zero-shot and many-shot (128 demonstrations) prompting. For regression-aware inference, we \new{adopt the best-performing variants from the original studies: sample-based RAIL \citep{Lukasik2024-yt} and grid-based RAFT \citep{Lukasik2025-ak}.} For predictive heads, we compare to linear regression (Lin.) and a two-layer MLP, with the MLP hidden size chosen to match {\method}’s trainable parameter count for each LLM backbone. 
\new{We adopt the Huber loss and y-normalization for predictive head baselines and {\method} only.} 
All baselines are evaluated on both frozen and LoRA fine-tuned LLM backbones, except for LoRA with many-shot prompting, since the benefits of in-context learning often diminish after fine-tuning \citep{yin-etal-2024-deeper, duan-etal-2024-exploring}.

We use $M=8$, $d_h = 256$, and $L=3$ for {\method}. See $\S$~\ref{sec:experimental_setup_extended} for additional details on {\method} ($\S$~\ref{sec:relish_setup_extended}), LLM backbones ($\S$~\ref{sec:llm_setup_extended}), datasets ($\S$~\ref{sec:datasets_setup_extended}), metrics ($\S$~\ref{sec:metrics_setup_extended}), and baselines ($\S$~\ref{sec:baseline_setup_extended}).

\begin{table}[t]
\centering
\small
\setlength{\tabcolsep}{0.8pt}
\renewcommand{\arraystretch}{1.3}

\begin{tabular*}{\columnwidth}{@{\extracolsep{\fill}} l | ccc | ccccccc | c @{}}
\toprule
& \multicolumn{3}{c|}{\textbf{Unsupervised}} & \multicolumn{7}{c|}{\textbf{Supervised}} & \multicolumn{1}{c}{\textbf{Ours}} \\
\cmidrule(r){2-4} \cmidrule(lr){5-11} \cmidrule(l){12-12}
\textbf{Metric} & \textbf{Zero} & \textbf{Many} & \textbf{RAIL} & \textbf{Lin.} & \textbf{MLP} & \textbf{L+Zero} & \textbf{L+Lin.} & \textbf{L+MLP} & \textbf{L+RAIL} & \textbf{RAFT} & \textbf{RELISH} \\
\midrule
\textbf{$r$} $\uparrow$ & $55.0_{\scriptscriptstyle \pm 0.3}$ & $52.3_{\scriptscriptstyle \pm 0.8}$ & $56.0_{\scriptscriptstyle \pm 0.1}$ & $44.4_{\scriptscriptstyle \pm 0.0}$ & $56.3_{\scriptscriptstyle \pm 0.4}$ & $57.2_{\scriptscriptstyle \pm 0.4}$ & $44.9_{\scriptscriptstyle \pm 0.0}$ & $55.9_{\scriptscriptstyle \pm 0.5}$ & $61.8_{\scriptscriptstyle \pm 0.5}$ & $66.1_{\scriptscriptstyle \pm 0.3}$ & \cellcolor{green!20} $\mathbf{72.9_{\scriptscriptstyle \pm 0.1}}$ \\
\textbf{$\rho$} $\uparrow$ & $55.5_{\scriptscriptstyle \pm 0.2}$ & $52.6_{\scriptscriptstyle \pm 1.0}$ & $58.8_{\scriptscriptstyle \pm 0.1}$ & $42.5_{\scriptscriptstyle \pm 0.0}$ & $53.7_{\scriptscriptstyle \pm 0.4}$ & $56.1_{\scriptscriptstyle \pm 0.5}$ & $43.0_{\scriptscriptstyle \pm 0.0}$ & $53.4_{\scriptscriptstyle \pm 0.3}$ & $62.9_{\scriptscriptstyle \pm 0.6}$ & $65.3_{\scriptscriptstyle \pm 0.5}$ & \cellcolor{green!20} $\mathbf{71.4_{\scriptscriptstyle \pm 0.3}}$ \\
\textbf{NRMSE} $\downarrow$ & $25.1_{\scriptscriptstyle \pm 0.0}$ & $22.2_{\scriptscriptstyle \pm 0.5}$ & $25.5_{\scriptscriptstyle \pm 0.0}$ & $23.1_{\scriptscriptstyle \pm 0.0}$ & $17.8_{\scriptscriptstyle \pm 0.3}$ & $18.9_{\scriptscriptstyle \pm 0.1}$ & $23.5_{\scriptscriptstyle \pm 0.2}$ & $17.7_{\scriptscriptstyle \pm 0.1}$ & $17.9_{\scriptscriptstyle \pm 0.1}$ & $15.7_{\scriptscriptstyle \pm 0.2}$ & \cellcolor{green!20} $\mathbf{12.6_{\scriptscriptstyle \pm 0.2}}$ \\
\bottomrule
\end{tabular*}
\caption{\new{Predictive performance macro-averaged across six datasets}, four LLMs, and three runs for each baseline. We report Pearson ($r$) and Spearman ($\rho$) correlations ($\uparrow$ higher is better) and range-Normalized Root Mean Squared Error (NRMSE) ($\downarrow$ lower is better), all scaled to \%. L+ denotes LoRA-based fine-tuning. We also report standard deviation ($\pm$).}
\label{tab:relish_overall_main_with_statllm}
\end{table}

\subsection{Results: Predictive Performance}
\label{sec:results}

\begin{figure}[ht]
    \centering
    \includegraphics[width=1\linewidth]{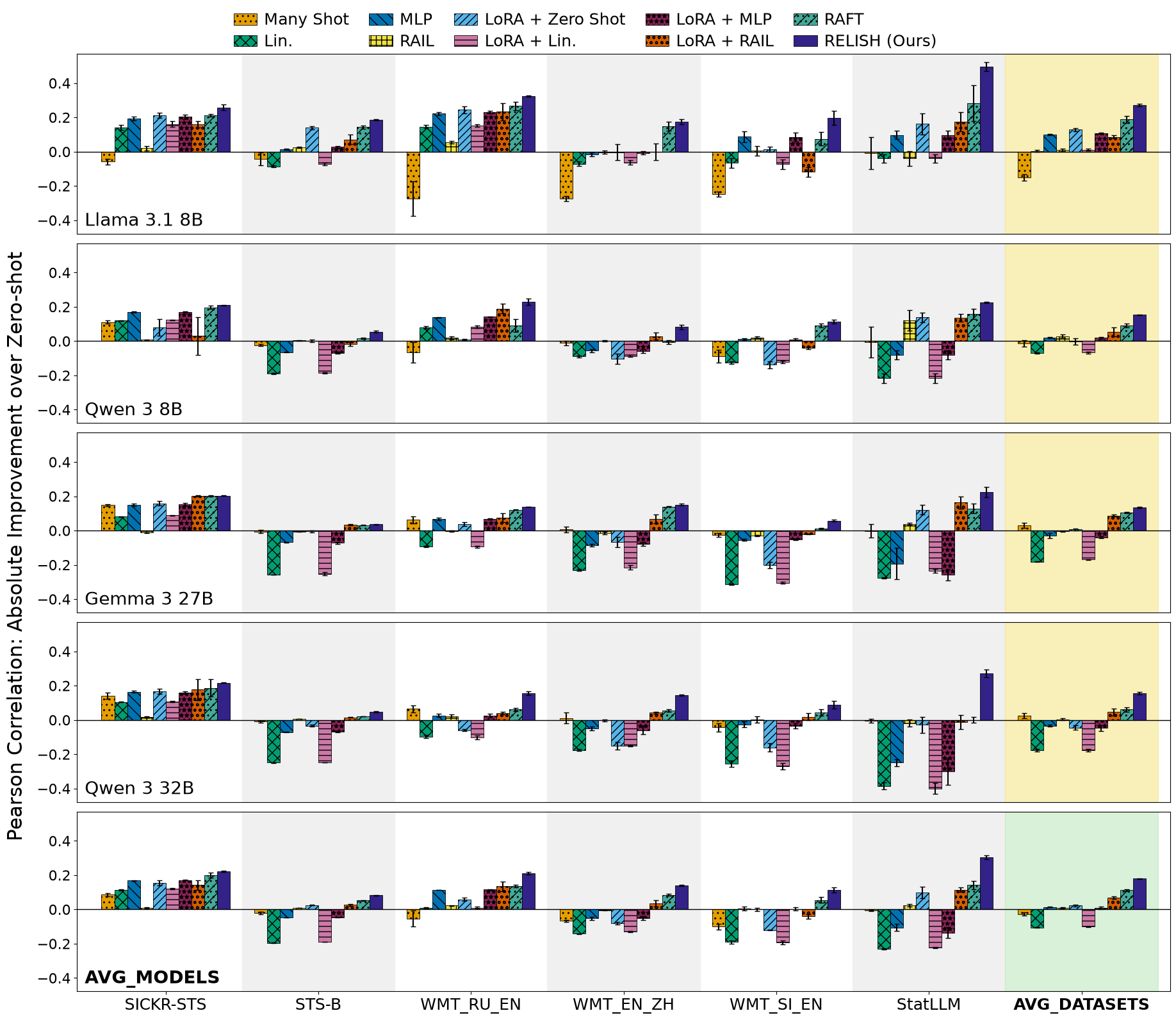}
\caption{\new{\label{fig:main_result_condensed}\textbf{Detailed results across datasets (columns) and LLMs (rows).}} The rightmost column averages over all datasets, and the bottom row averages over LLMs. Bars show absolute improvement in Pearson ($r$) correlation over the zero-shot baseline (the "ground zero" line). Error bars denote variability across 3 independently-seeded runs. All metrics are computed using gold test labels. 
{\method} consistently performs best across all settings.}
\end{figure}

\new{\textbf{Table~\ref{tab:relish_overall_main_with_statllm}} shows that {\method} consistently outperforms both unsupervised and supervised baselines across four LLM backbones and six datasets.} \textbf{Figure~\ref{fig:main_result_condensed}} shows detailed results for Pearson correlation relative to the zero-shot performance of the underlying LLM backbones. Across all \new{evaluated configurations}, {\method} consistently achieves the largest positive gains. In contrast, the other baselines are unreliable: they may yield comparable or modest improvements in some cases, but can degrade below zero-shot performance (negative bars).

Full predictive performance tables are reported in Tables~\ref{tab:stsb_relish_main}, \ref{tab:sickr_sts_relish_main}, \ref{tab:wmt_en_ru_relish_main}, \ref{tab:wmt_en_zh_relish_main}, \ref{tab:wmt_si_en_relish_main}, \new{and~\ref{tab:statllm_code_quality_relish_main}} (Appendix). There, we observe that {\method} achieves the highest predictive performance in almost all cases, except for a minor case in STS-B where Spearman is comparable to LoRA + RAIL \citep{Lukasik2024-yt} for Gemma 3 27B Instruct. See $\S$~\ref{sec:app_pred_perf} for extended discussions.

\new{We conduct ablations in \mbox{$\S$~\ref{sec:app_ablations}} to validate {\method}'s design choices. 
\mbox{$\S$~\ref{sec:app_ablations_loss}} compares {\method} trained with Huber and MSE, showing that gains are primarily architectural rather than loss-dependent and that {\method} still outperforms RAFT under squared-loss settings. 
\mbox{$\S$~\ref{sec:app_ablations_L}} ablates refinement depth $L$, suggesting that attention pooling alone ($L=1$) is strong but iterative updates ($L>1$) can improve predictive performance. 
\mbox{$\S$~\ref{sec:app_ablations_data}} and \mbox{$\S$~\ref{sec:app_ablations_runtime}} investigate the data efficiency of {\method} compared to other baselines. Results show that while RAFT performs best under scarce data regimes ($<$ 256 examples), {\method} scales better with moderate-to-large supervision ($>$ 512 examples) while being substantially faster to train than RAFT. 
\mbox{$\S$~\ref{sec:app_ablations_inference_time}} indicates that {\method} adds negligible inference overhead relative to simple predictive heads ($\sim$1\% slower), while being much faster than autoregressive and regression-aware baselines. 
Finally, \mbox{$\S$~\ref{sec:app_ablations_warm_start}} shows that input-dependent warm starts for $r^{(0)}$, such as last-token or mean-pooling, do not consistently lead to better predictive performance or faster convergence over the default random initialization.}
We discuss limitations in \mbox{$\S$~\ref{sec:app_limitations}}.

\subsection{Results: Training Footprint}
\label{sec:results_footprint}

\begin{table*}[ht]
\centering
\small 
\setlength{\tabcolsep}{10pt} 
\renewcommand{\arraystretch}{1.2}
\begin{tabular}{@{} l r r r r @{}}
\toprule
\textbf{Method} & \textbf{Llama 3.1 8B} & \textbf{Qwen 3 8B} & \textbf{Gemma 3 27B} & \textbf{Qwen 3 32B} \\
\midrule
LoRA / RAFT      & 21.0M (0.26\%) & 21.8M (0.27\%) & 116.5M (0.42\%) & 134.2M (0.41\%) \\
\textbf{{\method}} & \textbf{3.4M (0.04\%)} & \textbf{3.4M (0.04\%)} & \textbf{3.7M (0.01\%)} & \textbf{3.7M (0.01\%)} \\
\bottomrule
\end{tabular}
\caption[Trainable parameter sizes for different LLM backbones]{\textbf{How trainable parameter sizes vary for different LLM backbones.} We report the total trainable parameters and their \% relative to the underlying LLM. RAFT applies a regression-aware loss to LoRA adapters, sharing the same trainable footprint as LoRA with cross-entropy loss. While LoRA's budgets scale with model size, {\method} remains essentially invariant to the backbone, scaling instead with the hidden dimension (Table~\ref{tab:lightweight_detailed}).}
\label{tab:lightweight}
\end{table*}

{\method} trains a small set of modules atop a frozen LLM backbone, yielding a small trainable footprint. \textbf{Table~\ref{tab:lightweight}} shows that \textbf{{\method} consistently requires only \(\sim\)3.4--3.7M trainable parameters across all LLM backbones (0.01--0.04\%)}, whereas LoRA's trainable budget (and by extension RAFT's, which utilizes the same adapter architecture) grows with model size (0.26--0.42\%). Moreover, \textbf{this gap widens as LLM backbones scale.} Moving from 8B-class models to 27B--32B models increases LoRA parameters substantially, but {\method} remains nearly unchanged. Holding other parameters constant, \textbf{{\method}'s trainable components scale primarily with the backbone hidden dimension} (e.g., 4096 vs.\ 5120 units), rather than the total number of model parameters. See Table~\ref{tab:lightweight_detailed} and $\S$~\ref{sec:app_footprint} for details.

\section{Discussion}
\vspace{-0.5em}

\subsection{Broader Applications}

{\method} may also benefit a broader set of use cases, such as reward modeling (RM) \citep{ouyang2022training, wang2024secrets, xu-etal-2024-aligning, Liu2024-wi}, LLM-as-a-judge (LLMaaJ) \citep{zheng2023judging, li-etal-2025-generation, Chiang2025-ax}, and confidence estimation \citep{Kadavath2022-bf, Tian2023-ah, xiong2024can, Steyvers2025-no}. 
For example, TRACT \citep{Chiang2025-ax} fine-tunes LLMaaJ to output rubric-based scalar scores alongside chain-of-thought rationales under a regression-aware objective. \new{See $\S$~\ref{sec:app_cot_relish} for discussion of how {\method} could be extended with chain-of-thought (CoT) rationales.}

However, the supervised training underlying {\method} may represent a practical barrier for adoption in some settings. Many \textit{reference-free} LLMaaJ systems leverage black-box models that rely solely on prompt steering rather than supervised training \citep{li2024llms, li-etal-2025-generation}. Likewise, many confidence estimation methods obtain scores directly from black-box models through verbalization without supervision \citep{Kadavath2022-bf, Tian2023-ah, xiong2024can, Steyvers2025-no}. RM differs more fundamentally: its scalar rewards are typically learned from pairwise preference data and are only meaningful for comparing responses under the same prompt \citep{ouyang2022training, wang2024secrets}. We thus view these applications as potential extensions to this work.

\subsection{Point Prediction vs.\ Distribution Elicitation}
\label{sec:related:dist}

This work focuses on natural language regression, where the goal is to make a point estimate given textual inputs. A complementary line of work treats LLMs as probabilistic predictors and studies how to best elicit \emph{distributions} for uncertainty quantification \citep{Requeima2024-yg, Vedula2025-yn, Wang2025-bq, Hsu2026-bo, kambhatla-etal-2026-improving, piskorz2026eliciting}. While conceptually related, how best to elicit distributions from LLMs is beyond the scope of this work. Nonetheless, {\method} could yield distributions by replacing the linear regressor with quantile heads trained with pinball loss. 



\section{Conclusion}
\label{sec:conclusion}
\vspace{-0.5em}

We proposed {\method}, a novel, lightweight architecture for LLM-based natural language regression that iteratively refines a learned latent state via cross-attention. Across \new{six} datasets and four LLM backbones, {\method} consistently outperforms prior state-of-the-art regression baselines. In addition, {\method} requires only $\sim$3.4--3.7M trainable parameters (0.01--0.04\% additional overhead) and avoids LoRA scaling bottlenecks. Promising applications include uncertainty quantification, reward modeling, and LLM-as-a-judge tasks.


\section*{\new{Acknowledgments}}

We appreciate the valuable feedback we received on this manuscript from our anonymous reviewers, Junyi Jessy Li, Leqi Liu, and members of the Artificial Intelligence and Human-Centered Computing (AI\&HCC) lab at UT Austin. This work was supported by the NSF under Cooperative Agreement 2421782 and the Simons Foundation grant MPS-AI-00010515, awarded to the NSF-Simons AI Institute for Cosmic Origins (CosmicAI)\footnote{\url{https://www.cosmicai.org/}}. This research was also supported by computational resources provided by the Texas Advanced Computing Center (TACC) at The University of Texas at Austin. The statements made herein are solely the opinions of the authors and do not reflect the views of the sponsoring agencies.




\bibliography{colm2026_conference, paperpile}
\bibliographystyle{colm2026_conference}

\clearpage

\appendix

\section{Appendix Organization}
\label{sec:app_roadmap}

We organize the Appendix as follows: 

\begin{itemize}
    \item \new{\textbf{Extended Background ($\S$~\ref{sec:app_expanded_related})} expands related works ($\S$~\ref{sec:background}) on clarifying the operational costs of sample- and grid-based RAIL/RAFT ($\S$~\ref{sec:app_rail_raft_costs}), conceptual limitations of RAFT ($\S$~\ref{sec:app_relish_vs_raft}), and why {\method} is preferable to \texttt{[CLS]} pooling ($\S$~\ref{sec:app_cls_pooling}).}
    
    \item \textbf{Experimental Setup ($\S$~\ref{sec:experimental_setup_extended})} complements $\S$~\ref{sec:experimental_setup} with granular details on {\method} ($\S$~\ref{sec:relish_setup_extended}), LLM backbones ($\S$~\ref{sec:llm_setup_extended}), datasets ($\S$~\ref{sec:datasets_setup_extended}), metrics ($\S$~\ref{sec:metrics_setup_extended}), and baselines ($\S$~\ref{sec:baseline_setup_extended}).
    
    \item \textbf{Predictive Performance ($\S$~\ref{sec:app_pred_perf})} expands on the main results ($\S$~\ref{sec:results}) by detailing the metrics aggregation pipeline ($\S$~\ref{sec:app_agg_pipeline}), providing additional tables and figures ($\S$~\ref{sec:app_pred_perf_figures_and_tables}), and analyzing both high-level ($\S$~\ref{sec:app_pred_perf_patterns_global}) and dataset-specific ($\S$~\ref{sec:app_pred_perf_patterns_local}) patterns \new{across STS, WMT, and code quality estimation}.

    \item \textbf{Training Footprint ($\S$~\ref{sec:app_footprint})} expands on the parameter efficiency and training footprint discussion from $\S$~\ref{sec:results_footprint}.
    
    \item \textbf{Ablation Studies ($\S$~\ref{sec:app_ablations})} complements the main predictive performance results ($\S$~\ref{sec:results}) by comparing {\method} trained with Huber versus MSE loss ($\S$~\ref{sec:app_ablations_loss}), examining the model's sensitivity to the refinement depth $L$ ($\S$~\ref{sec:app_ablations_L}), \new{and evaluating supervised-data scaling ($\S$~\ref{sec:app_ablations_data}), training runtime ($\S$~\ref{sec:app_ablations_runtime}), inference wall-clock throughput ($\S$~\ref{sec:app_ablations_inference_time}), and warm-start initialization ($\S$~\ref{sec:app_ablations_warm_start}).}
    
    \item \textbf{Limitations ($\S$~\ref{sec:app_limitations})} concludes with a discussion of the method's limitations and avenues for future work, \new{including CoT ($\S$~\ref{sec:app_cot_relish}) and RAFT-style hybridization ($\S$~\ref{sec:app_relish_raft_hybrid}) as future extensions.}
\end{itemize}




\section[Expanded Related Work and Conceptual Comparisons]{\new{Expanded Related Work and Conceptual Comparisons}}
\label{sec:app_expanded_related}

\new{
In this section, we provide additional details on how {\method} differentiates from prior LLM-based regression approaches. We first clarify the operational costs of sample- and grid-based RAIL/RAFT ($\S$~\ref{sec:app_rail_raft_costs}), then discuss conceptual limitations of RAFT ($\S$~\ref{sec:app_relish_vs_raft}) and why {\method} is preferable to \texttt{[CLS]} pooling ($\S$~\ref{sec:app_cls_pooling}).
}

\subsection[Operational Costs of Sample- and Grid-based RAIL/RAFT]{\new{Operational Costs of Sample- and Grid-based RAIL/RAFT}}
\label{sec:app_rail_raft_costs}

\new{
To compute Bayes-optimal statistics, regression-aware approaches such as RAIL \citep{Lukasik2024-yt} and RAFT \citep{Lukasik2025-ak} must estimate the posterior of the LLM's induced output distribution through \textit{multiple samples} or \textit{enumerating a grid of candidates} (e.g., {`1', `2', `3', `4', `5'}). It follows that the operational costs of sample-based variants scale linearly with the number of samples, and are generally more expensive compared to one pass autoregressive methods and predictive heads. 
}

\new{
For grid-based RAIL and RAFT, \citet{Lukasik2025-ak} demonstrate (see their Table 2) that when grid candidates are single tokens (e.g., `0' or `1'), the computation simplifies to extracting the next-token probability, implying that inference overhead may be minimal for ordinal regression contexts. However, our setting uses continuous targets serialized to three decimal places (e.g., `1.250'), which are typically multi-token strings. Here, the probability of generating a multi-token candidate must be computed autoregressively as the product of conditional token probabilities:
\[
\new{p(\mathrm{str}(y)\mid x) = \prod_{j=1}^{|\mathrm{str}(y)|} p(t_j \mid x, t_{<j}).}
\]
It follows that the operational cost of grid-based variants scales both with the grid size and tokenized length of the numeric string targets. Since RAFT also fine-tunes the LLM backbone, these per-candidate probabilities must also remain in the computational graph for backpropagation, thereby imposing additional memory usage. In contrast, {\method} directly predicts a scalar from a single forward pass using a lightweight predictive head. 
}

\subsection[Why is {\method} Conceptually Preferable to RAFT?]{\new{Why is {\method} Conceptually Preferable to RAFT?}}
\label{sec:app_relish_vs_raft}

\new{Beyond efficiency, {\method} offers conceptual advantages over RAFT \citep{Lukasik2025-ak} by directly mapping text to scalars, whereas RAFT indirectly derives a prediction from probability distributions over textualized scalars. Specifically, RAFT computes a weighted sum over a grid of numeric string candidates. Thus, RAFT is structurally constrained by the modality mismatch between tokens and scalars. This is problematic for several reasons.
}

\new{First, RAFT is prone to the \textbf{verbalization gap}, where LLMs internally encode numerical magnitudes with high precision but fail to express this knowledge through generated tokens \citep{yuchi-du-eisner-2026}. Abstractly, RAFT can be viewed as calibrating the LLM's ability to assign precise relative probability mass (logits) to textualized numeric targets in order to ``triangulate" an accurate weighted sum. Thus, if the model's latent numerical reasoning is not perfectly translated to verbalized logits, then the resulting expectation becomes inherently noisy.}

\new{As an example of the verbalization gap, RAFT is prone to \textbf{probability leakage}. As noted in Equation 10 of \citet{Lukasik2025-ak}, RAFT derives weights from the probability distribution over the \textit{entire} vocabulary, not just the grid candidates. Thus, if the LLM is uncertain, with probability mass distributed across both candidate and non-candidate tokens, the total weights for the grid targets may not sum to 1. This means that the weighted sum can artificially deflate toward zero, potentially biasing lower-magnitude predictions. Consequently, RAFT may conflate numerical magnitude with model uncertainty, since it is hard to discern whether a low-value output is genuine or merely a sign of probability leakage in verbalization.}

\new{Second, this possibility for probability leakage can lead to \textit{asymmetrical bounding and inference instability}. While Lemma 3 in \citet{Lukasik2025-ak} proves that the model can represent any target in the grid interval $[a,b]$ via a convex combination, this is not guaranteed under leakage, when the grid probabilities sum to less than 1. Effectively, RAFT may emit a prediction below the lower endpoint through verbalization failure rather than genuine extrapolation. Furthermore, RAFT is mathematically constrained to predict values no greater than the upper endpoint of the grid. Thus, RAFT cannot generalize to out-of-distribution magnitudes or unbounded continuous targets.}

\new{Trying to mitigate these generalization limitations is non-trivial, since simply increasing the grid range can introduce instability. Per Lemma 2 in \citet{Lukasik2025-ak}, a small $ \epsilon$-shift in the probability distribution can result in a lower-bound prediction error of $\geq \frac{b}{4}^2 \epsilon$. Consequently, for large grid candidates, minor miscalibrations at inference time can drastically skew the final prediction.}

\new{Third, the verbalization gap is exacerbated by \textit{tokenization fragility}. As shown by \citet{wallace-etal-2019-numbers} and \citet{geva-etal-2020-injecting}, transformers often employ subword tokenizers such as BPE variants \citep{sennrich-etal-2016-neural} that fracture numbers into token fragments, disrupting numerical continuity. Since RAFT requires computing the conditional probability of generating textualized numeric targets, the method may be sensitive to superficial formatting. For instance, the probability of generating ``1'' versus ``1.000'' might differ substantially, since the LLM must aggregate probability mass across multiple fragments to generate the latter. Thus, calibrating logits for multi-token grid candidates can be unstable.}

\new{Finally, RAFT may be brittle due to \textit{semantic entanglement} from pre-training, which fine-tuning may not fully overcome. For instance, how an LLM encodes ``5'' is highly contextual (e.g., ``5 o'clock'' versus ``5\%''). As a result, token representations may be pulled in multiple semantic directions during fine-tuning. This is supported by the performance degradation observed in \citet{Lukasik2025-ak} when numeric grid candidates are replaced with their date equivalents (e.g., using ``January'' for ``1''). Forcing the LLM to represent grid candidates purely as continuous ordinal values can therefore make the resulting probability distribution highly brittle.}

\new{{\method} bypasses the above limitations by avoiding token probabilities entirely. Instead, {\method} directly maps LLM hidden states to a continuous scalar, thereby avoiding the verbalization gap and probability leakage. Likewise, {\method} avoids tokenization fragility and semantic entanglement by avoiding textualized numeric targets altogether. While RAFT is mathematically bounded by its grid candidates and potentially fragile for wide intervals, {\method} predicts an unbounded standardized scalar through a continuous predictive head.}

\subsection[Why Prefer RELISH over CLS Pooling?]{\new{Why Prefer {\method} over \texttt{[CLS]} Pooling?}}
\label{sec:app_cls_pooling}

\new{{\method} is preferable over \texttt{[CLS]} pooling for two reasons. First, many modern LLMs are decoder-only and thus lack a native \texttt{[CLS]} token. In encoder-only models like BERT, \texttt{[CLS]} is explicitly learned during pre-training, since the model uses the \texttt{[CLS]} token representation to predict whether two input segments are consecutive sentences \citep{Devlin2019-qp}. In contrast, decoder-only LLMs are trained to generate tokens autoregressively. Therefore, they do not need to learn an analogous \texttt{[CLS]}-like token, since there is no sequence-level pre-training objective that uses it.}

\new{Second, {\method} avoids any structural overhead introduced by \texttt{[CLS]} pooling. Specifically, a \texttt{[CLS]}-like solution for decoder-only LLMs would require either adding a new vocabulary token or selecting a special unused token, if any, for further training, so that this token learns how to attend to other tokens. If the token is new, the tokenizer and embedding layers would also need to be updated.} 

\new{In contrast, {\method} offers a self-contained module that separates representation learning from predictive forecasting. {\method} does not modify the LLM backbone in any way. Instead, {\method} learns a standalone predictive head that operates on frozen LLM token representations to make accurate predictions. This modularity confines task-specific fine-tuning to the predictive head and allows the same architectural design to generalize to different model backbones.}

\new{For decoder-only LLMs, a cheap \texttt{[CLS]}-like solution might be to use the representation of the last token, such as \texttt{[EOS]}. We do not benchmark against this approach because \citet{tang2025understanding} show that predictive heads using last-token representations perform worse than those using mean pooling. \citet{Lukasik2025-ak} similarly show that RAFT outperforms \texttt{[CLS]} pooling predictive heads. Since {\method} outperforms RAFT in our experiments, we focus our comparisons on stronger baselines rather than ablating pooling variants.}

\section{Expanded Experimental Setup}
\label{sec:experimental_setup_extended}

In this section, we provide further details of the experimental setup discussed in $\S$~\ref{sec:experimental_setup} on {\method} ($\S$~\ref{sec:relish_setup_extended}), LLM backbones ($\S$~\ref{sec:llm_setup_extended}), datasets ($\S$~\ref{sec:datasets_setup_extended}), metrics ($\S$~\ref{sec:metrics_setup_extended}) and baselines ($\S$~\ref{sec:baseline_setup_extended}).

To ensure statistical robustness and reproducibility of our empirical results, we conduct three independent runs of experiments using three different random seeds (\textsc{42, 1234}, and \textsc{2026}) to initialize pseudo-random number generation. We then report aggregated metrics across trials to account for this natural variance. Table~\ref{tab:hyperparams} summarizes all hyperparameters.

\subsection{\method}
\label{sec:relish_setup_extended}

For {\method}, we keep the LLM backbone frozen and train only the projection matrix $W \in \mathbb{R}^{d \times d_h}$, the latent state $r^{(0)}$, the attention module, and the linear output linear. We use a learning rate of $1e-4$, eight attention heads, a hidden size of $d_h=256$, and a Huber $\delta = 1.0$ for all datasets and LLM backbones. Our FFN residual MLP consists of two linear layers with a hidden size of $4 * d_h$ interpolated by a GELU \citep{hendrycks2016gaussian}. See Table~\ref{tab:hyperparams} for all hyperparameters. 

\paragraph{What if the initial projection is omitted?}

In $\S$~\ref{sec:method}, we noted that the initial projection from the frozen LLM backbone states $H(x) \in \mathbb{R}^{S \times d}$ into $X(x) \in \mathbb{R}^{S \times d_h}$ using $W \in\mathbb{R}^{d\times d_h}$ is mathematically optional whenever the chosen head dimension matches the backbone's hidden dimension ($d_h = d$). If this optional step is not taken, the token-level memory directly utilizes the raw LLM hidden states (i.e., $X(x) = H(x)$). 

\textit{Omitting this initial projection is highly parameter-inefficient}. By directly utilizing the raw hidden states ($X(x) = H(x)$), the model bypasses the projection matrix $W$, reducing the total parameter count by $d^2$ initially. However, the $L$ downstream refinement layers now operate at the full LLM backbone's dimension $d$ instead of $d_h$. Within each refinement layer, each multi-head cross-attention block introduces at least $\approx 3d^2$ additional parameters, since each projection matrix for the query, value, and key ($W_m^Q, W_m^K, W_m^V$) now operate in $d$. Similarly, each feed-forward network introduces $\approx 8d^2$ additional parameters, assuming our 2-layer setup where each layer is now of shape $d \times 4d$. Thus, omitting the initial projection at least contributes $\mathcal{O}(L \cdot 11d^2)$ additional parameters. Consequently, the $d^2$ parameters saved by dropping $W$ are vastly overshadowed by the quadratic parameter explosion in the refinement steps. 

Empirically, this quadratic scaling no longer renders {\method} as lightweight. Under our configuration with $L=3$, setting $d_h=d$ scales the trainable footprint to approximately $32d^2$. For our LLM backbones with large hidden state dimensions (e.g., $d \ge 4096$ versus $d_h =256$, see Table~\ref{tab:lightweight_detailed}), this effectively scales overall parameters by a factor of $\sim 512$, requiring over hundreds of millions of parameters. As a result, this full-dimension configuration introduces \textit{substantially more trainable parameters} than even LoRA. We suspect that this quadratic scaling will yield diminishing returns in predictive performance that will not justify this explosive additional overhead. 

Moreover, beyond the severe parameter inefficiency, omitting this projection step forces the iterative refinement mechanism to operate directly on the raw pre-trained representation space, prohibiting the model's ability to transform those frozen representations into a specialized, task-specific manifold before the cross-attention interactions begin. Therefore, we do not experiment with setting $d_h = d$ in our experiments. 

\paragraph{\new{Why the Penultimate Layer?}}
\label{sec:app_penultimate_layer}

\new{Our design to extract representations from the penultimate layer (directly before the language modeling head) follows standard practice in LLM-based regression. \citet{tang2025understanding} fit an MLP head on mean-pooled token representations after a full forward pass of a transformer. Likewise, \citet{Lukasik2025-ak} benchmark RAFT against predictive head baselines that mean-pool output embeddings from the penultimate layer, following RankT5 \citep{Zhuang2023-om}. Traditionally, canonical embedding models such as Sentence BERT \citep{reimers-gurevych-2019-sentence} emit a vector representation by pooling hidden-layer representations. In reward modeling, perhaps the most prominent application of LLM-based regression, \citet{zhu2023principled} note that features are usually derived from LLMs by removing the language-modeling head and attaching a linear scalar head directly to the final-layer hidden states. Later reward modeling systems adopt a similar formulation \citep{miao2025energy, li2026provably}.}

\new{Nevertheless, the optimal layer for regression may not always be the penultimate layer. Different LLMs naturally differ in architecture, including depth, backbone structure, and attention mechanism, so intermediate layers may serve different functions across models \citep{Zhang2024-sz, skean2025layer, NEURIPS2025_eae3af0f}. Consequently, the best layer or layer combination is likely model-specific. In contrast, the penultimate layer is a robust, model-agnostic choice because it is the definitive stop before the language-modeling head across architectures. Therefore, we do not investigate other layers or weighted layer combinations in this study.}

\subsection{LLM Backbones}
\label{sec:llm_setup_extended}

We source our LLM backbones directly from public Huggingface checkpoints: \textbf{Llama-3.1-8B-Instruct},\footnote{\url{https://huggingface.co/meta-llama/Llama-3.1-8B-Instruct}} \textbf{Qwen 3-8B},\footnote{\url{https://huggingface.co/Qwen/Qwen 3-8B}} \textbf{Qwen 3-32B},\footnote{\url{https://huggingface.co/Qwen/Qwen 3-32B}} and \textbf{Gemma-3-27B-Instruct}.\footnote{\url{https://huggingface.co/google/gemma-3-27b-it}}.

\begin{table}[t]
\centering
\small
\begin{tabular}{l r r r c}
\toprule
\textbf{Dataset Name} & \textbf{Train} & \textbf{Validation} & \textbf{Test} & \textbf{Score Range} \\
\midrule
STS-B & \new{5{,}749} & 1{,}500 & 1{,}380 & [0, 5] \\
SICKR-STS             & 6,948     & 1,985        & 994 & [1, 5] \\
WMT\_EN\_ZH                & 7{,}000 & 1{,}000   & 1{,}000 & [0, 100] \\
WMT\_RU\_EN                & 7{,}000 & 1{,}000   & 1{,}000 & [0, 100] \\
WMT\_SI\_EN                & 7{,}000 & 1{,}000   & 1{,}000 & [0, 100] \\
\new{StatLLM} & \new{321} & \new{100} & \new{200} & \new{[0, 50]} \\
\bottomrule
\end{tabular}
\caption{Dataset split statistics.}
\label{tab:datasets}
\end{table}

\subsection{Datasets}
\label{sec:datasets_setup_extended}

We further provide details on the origin, composition, annotation, and post-processing of the \new{six} natural language regression datasets used in our experiments. Table~\ref{tab:datasets} summarizes the splits we adopt for each dataset.

\paragraph{STS-Benchmark (STS-B)} STS-B\footnote{\url{https://huggingface.co/datasets/mteb/stsbenchmark-sts}} is a compilation of the English datasets used in the SemEval STS shared tasks from 2012 to 2017. \citet{cer-etal-2017-semeval} curated STS-B to provide a standard benchmark for the STS community. Sentence pairs are sampled from three distinct domains: image captions, video captions, and news headlines. Annotations were collected via crowdsourcing (Amazon Mechanical Turk), where workers assigned scores from 0 to 5 based on the degree of semantic equivalence. We followed the split created by \citet{enevoldsen2025mmteb}. 

\paragraph{SICKR-STS} Derived from the Sentences Involving Compositional Knowledge (SICK) dataset \citep{marelli-etal-2014-sick}, SICKR-STS\footnote{\url{https://huggingface.co/datasets/mteb/sickr-sts}} focuses specifically on semantic relatedness. The sentences originated from the 8K ImageFlickr \citep{hodosh2013framing} and MSR-Video Description \citep{chen-dolan-2011-collecting} datasets and were subsequently processed to include specific linguistic variations, such as negation and passive-to-active voice transformations. Like STS-B, it was annotated by crowdsourced workers on a 5-point Likert scale. We followed the split created by \citet{enevoldsen2025mmteb}.

\paragraph{WMT2020 Multilingual Quality Estimation (MLQE) Datasets} The MLQE\footnote{\url{https://huggingface.co/datasets/wmt/wmt20_mlqe_task1}} ``Task 1: Predicting sentence-level DA'' involves predicting the quality of neural machine translation (NMT) outputs, without access to human references. We select three subsets to test model performance across varying data availability: English-to-Chinese (En–Zh; high-resource), Russian-to-English (Ru–En; medium-resource), and Sinhala-to-English (Si–En; low-resource). The source sentences for En–Zh and Si–En are primarily from Wikipedia, while Ru–En includes a mix of Wikipedia and Reddit content. Each machine translation was directly assessed (DA) by professional translators who scored the output on a fine-grained, ordinal 0-100 scale based on its accuracy relative to the source sentence. We adopt the splits created by \citet{specia-etal-2020-findings-wmt} directly.

\paragraph{\new{StatLLM}} 

\new{
StatLLM \citep{Song2026StatLLM} evaluates the quality of LLM-generated SAS code for scientific tasks. While STS and WMT are canonical tasks in text regression, StatLLM comprises 207 scientific workflow tasks, each paired with code generated by GPT-3.5 \citep{openai2022chatgpt}, GPT-4 \citep{openai2023gpt4}, and Llama 3.1 70B \citep{grattafiori2024llama}, yielding a total of $207*3=621$ examples. This is notably smaller than the STS and WMT datasets, but code generation is also a more complex task, and examples are generally longer than those in STS and WMT. Also, whereas STS and WMT inputs are scored against references, each input in StatLLM is a single generated SAS program plus task instructions. StatLLM thus provides a different form of regression task for evaluating the generalization of methods.}

\new{Our interest in StatLLM is further motivated by emerging needs of AI-for-science (AI4Science) \citep{zheng-etal-2025-automation} where LLM use is becoming increasingly prevalent across a variety of use cases: to generate research ideas \citep{liu2026owns, GautamIdeate2026}, synthesize literature \citep{ma-etal-2026-intragent}, and run computational experiments \citep{qi-etal-2025-metascientist}. While existing AI4Science benchmarks primarily evaluate whether LLM-generated artifacts produce correct outputs \citep{ScienceAgentBench2025, astrovisbench2025}, estimating the quality of analytical code at runtime remains largely underexplored. Such inference-time quality estimates can assist scientists or autonomous agents in scientific workflows and to triage generated code before execution, flagging problematic snippets for human review \citep{Uncertain2026} or routing outputs back to an LLM for further revision \citep{hao2026racer}.}

\new{
Each example in StatLLM is evaluated by human experts along ten rubric criteria grouped under three dimensions: readability, executability, and correctness. The underlying rubric criteria use a five-point Likert scale, ranging from 1 (strongly disagree) to 5 (strongly agree). The  dataset\footnote{\url{https://github.com/yili-hong/StatLLM}} does not provide per-criterion ratings but only aggregated scores. Readability aggregates five criteria (5--25), executability aggregates two criteria (0--10), and correctness aggregates three criteria (0--15). The overall quality score is the sum of these three aggregate scores, yielding a final prediction target in the range 0--50. In our experiments, we directly predict the overall quality score without querying for intermediate dimensions. 
}

\subsection{Metrics}
\label{sec:metrics_setup_extended}

\paragraph{Range-Normalized Root Mean Square Error}
Because our datasets use heterogeneous label scales (e.g., STS-B: 0–5, SICK-R: 1–5, WMT: 0–100\new{, StatLLM: 0--50}), raw RMSE is not directly comparable across datasets. In addition, direct macro averaging over raw RMSE can disproportionately upweight the translation datasets, since the prediction errors are generally larger for the WMT task. In Table~\ref{tab:relish_overall_main_with_statllm}, we therefore report range-normalized RMSE (NRMSE), defined as $\mathrm{NRMSE} = \mathrm{RMSE} / (y_{\max} - y_{\min})$, or dividing the raw RMSE by each dataset’s gold-score range. We then macro-average NRMSE across datasets, models, and runs. This yields an error metric on a common unitless scale, where lower values indicate smaller prediction error relative to the particular dataset’s full scoring range.


\subsection{Baselines}
\label{sec:baseline_setup_extended}

This section provides further implementation details for the baseline methods (Table~\ref{tab:operational_tradeoffs}) benchmarked in our work, including specific hyperparameter configurations for both the frozen and LoRA-fine-tuned regimes.

\subsubsection{Autoregressive Decoding}

\paragraph{Zero-shot Prompting} 
We evaluate the base capabilities of the LLM backbones using the system prompts and formatting templates detailed in Table~\ref{tab:prompt_stsb}, ~\ref{tab:prompt_sickr}, ~\ref{tab:prompt_wmt_enzh}, ~\ref{tab:prompt_wmt_ruen}, ~\ref{tab:prompt_wmt_sien}\new{, and~\ref{tab:prompt_statllm}}. We limit model outputs to a maximum of 10 new tokens. We set temperature = 0.7 across three independent seed runs to control for output variance. The numeric score is directly extracted from the generated text using Python's \textsc{float} function. Note that this means the LLM could generate potentially multiple equivalent string representations of the same numeric candidate, such as `1.0' versus `1.00'. Since we employed only instruction-following models, all LLM backbones produced valid numeric strings.  

\paragraph{Many-shot Prompting} 
As shown in \citet{vacareanu2024from}, LLMs can be capable regressors when given in-context demonstrations. We therefore also experiment with many-shot learning in addition to zero-shot prompting. We provide $k=128$ in-context demonstrations for all datasets, which already incurs a $\approx 10\times$ slowdown in inference across all LLM backbones, so we did not experiment with larger numbers of demonstrations. Examples are randomly sampled from the training split for each independent run, but are consistent for trials using the same random seed. We use the same formatting instructions as the zero-shot case, prefixing the user message with formatted examples before appending the test query. As noted in the main body, we do not evaluate many-shot in the LoRA fine-tuning regime because the parameter updates during fine-tuning are intended to internalize the task logic, rendering in-context demonstrations (which were already seen during training) redundant. 

\subsubsection{Regression-Aware Baselines}

\paragraph{RAIL} 
Regression-aware inference for LLMs (RAIL) \citep{Lukasik2024-yt} avoids text decoding by treating the numeric prediction as a decision-theoretic problem. We implement the RAIL posterior mean approach, where the model's output distribution over numeric tokens is used to calculate an expected value. We limit the vocabulary to digits 0--9 and the decimal point, normalizing the probabilities over these tokens at each step of a fixed-length numeric sequence generation. Following \citet{Lukasik2024-yt}, we sampled 16 outputs from the LLM backbone to approximate the posterior with an effective temperature of $0.25$. 

\paragraph{RAFT} 
Regression-aware fine-tuning (RAFT) \citep{Lukasik2025-ak} extends RAIL by fine-tuning the model to minimize a special regression-aware loss (see Table~\ref{tab:regression-taxonomy}) rather than the standard token-level cross-entropy. In our experiments, RAFT uses the same hyperparameter setup as the standard LoRA decoding baselines, but optimizes the numeric expected value computed via the RAIL procedure using mean squared error. Following \citet{Lukasik2025-ak}, we do not sample like RAIL but instead pick $\mathcal{Y}_{\text{grid}}$ values that correspond to the gold-score range for each dataset. For STS-B, we use $\mathcal{Y}_{\text{grid}} = \{ 0, 1, 2, 3, 4, 5\}$. For SICKR-STS, we use $\mathcal{Y}_{\text{grid}} = \{ 1, 2, 3, 4, 5\}$. For WMT datasets, we use $\mathcal{Y}_{\text{grid}} = \{ 0, 10, 20, 30, 40, 50, 60, 70, 80, 90, 100\}$. \new{For StatLLM, we use $\mathcal{Y}_{\text{grid}} = \{ 0, 5, 10, 15, 20, 25, 30, 35, 40, 45, 50\}$.} \new{Following \citet{Lukasik2025-ak}, we do not normalize grid candidates for RAFT.} We observe that all LLM backbones produced continuous scalars and achieved good predictive performance even with only a few grid candidates. 

\subsubsection{Predictive Head Baselines}
\label{sec:app_pred_head}

\paragraph{Linear Regression (Lin.)}
We first extract token hidden states from the penultimate layer of the transformer,
$H(x)\in\mathbb{R}^{S\times d}$, and an attention mask
$m\in\{0,1\}^S$. We then compute the masked mean-pooled representation:
\[
\bar h(x)=\frac{\sum_{t=1}^S m_t\,H_t(x)}{\sum_{t=1}^S m_t}.
\]
Finally, we apply a single linear layer to predict the (standardized) scalar target:
\[
\hat z = \mathbf{w}^\top \bar h(x) + b.
\]

We omit the y-normalization and de-normalization specified in Section~\ref{sec:method} for brevity.

\paragraph{Multi-Layer Perceptron (MLP)}
We similarly first compute $\bar h(x)$ for the MLP predictive head. Following \citet{tang2025understanding}, we use a 2-layer MLP regressor
\[
u_1=\mathrm{Drop}\!\left(\mathrm{RELU}(W_1\bar h+b_1)\right),\quad
u_2=\mathrm{Drop}\!\left(\mathrm{RELU}(W_2u_1+b_2)\right),\quad
\hat z=W_3u_2+b_3,
\]
\[
W_1\in\mathbb{R}^{d\times h},\quad
W_2\in\mathbb{R}^{h\times h},\quad
W_3\in\mathbb{R}^{h\times 1}, \\
\]
with RELU activations \citep{glorot2011deep} and dropout rate $0.1$ after each hidden layer. To approximately match the training capacity of {\method}, we select hidden size $h$ by parameter matching. This yields $h=704$ for Llama 3.1 8B Instruct and Qwen 3 8B, and $h=640$ for Gemma 3 27B Instruct and Qwen 3 32B. We fine-tune both the linear regressor head and the 2-layer MLP using Huber loss with $\delta = 1.0$ and a learning rate of $1e-4$. See Table~\ref{tab:hyperparams} for all hyperparameters.

\subsubsection{LLM Training Configurations}

All LLM backbones were trained using the AdamW optimizer with a linear learning rate scheduler and a warmup phase of 10\% of the total training steps. LLM training was conducted for 5 epochs with an effective batch size of 32 and a learning rate of $5e-5$. We stop the training early after 2 epochs if the validation RMSE does not improve. 

\paragraph{Frozen Regime} 
In the frozen regime, the LLM backbone parameters $\theta$ are fixed. For RELISH, Linear, and MLP baselines, only the weights of the predictive heads are updated. We used a learning rate of $1e-4$, the Huber loss, and a maximum of 10 epochs, with a patience of 2 epochs to improve the validation loss. 

\paragraph{LoRA Regime} 
For fine-tuning, we apply Low-Rank Adaptation \citep{hu2022lora} to all linear projections within the transformer blocks: the attention modules (\texttt{q\_proj, k\_proj, v\_proj, o\_proj}) and the feed-forward network modules (\texttt{gate\_proj, up\_proj, down\_proj}). We use a learning rate of $5 \times 10^{-5}$. Detailed hyperparameters are provided in Table~\ref{tab:hyperparams}. \new{For all LoRA variants besides RAFT, we adopt the standard cross entropy loss on textualized targets, so we do not apply y-normalization here.}

\begin{table}
\centering
\small
\begin{tabular}{ll}
\toprule
\textbf{Hyperparameter} & \textbf{Value} \\ \midrule
\rowcolor[HTML]{F2F2F2} \multicolumn{2}{l}{\textit{Shared}} \\
Seed Values & 42, 1234, 2026 \\
Effective Batch Size & 32 \\
Early Stopping Patience & 2 \\
Warmup Ratio & 0.1 \\
Weight Decay & 0.01 \\
Dropout & 0.1 \\
Precision & bfloat16 \\ \midrule
\rowcolor[HTML]{F2F2F2} \multicolumn{2}{l}{\textit{LoRA / RAFT}} \\
Learning Rate & $5 \times 10^{-5}$ \\
Max Epochs & 5 \\
Rank ($r$) & 8 \\
Alpha ($\alpha$) & 16 \\
Target Modules & All Linear Layers (\texttt{q, k, v, o, gate, up, down}) \\ \midrule
\rowcolor[HTML]{F2F2F2} \multicolumn{2}{l}{\textit{Predictive Heads / RELISH}} \\
Learning Rate & $1 \times 10^{-4}$ \\
Max Epochs & 10 \\
\method{} $d_h$ & 256 \\
Predictive Head $d_h$ & 704 (Llama \& Qwen-8B) or 640 (Gemma \& Qwen-32B); see $\S$~\ref{sec:app_pred_head} \\
$L$ & 3 \\
FFN Hidden Size & 1024 \\ 
Huber $\delta$ & 1.0 \\ \bottomrule
\end{tabular}
\caption{Hyperparameter configurations.}
\label{tab:hyperparams}
\end{table}

\section{Full Predictive Performance Results}
\label{sec:app_pred_perf}

In this section, we expand on the predictive performance results discussed in $\S$~\ref{sec:results}. First, we discuss how we aggregated results across datasets, LLMs, and seeds in Table~\ref{tab:relish_overall_main_with_statllm}. We then introduce additional figures and tables ($\S$~\ref{sec:app_pred_perf_figures_and_tables}). Next, we discuss the high-level trends and provide some intuition on why {\method} works so well ($\S$~\ref{sec:app_pred_perf_patterns_global}). Finally, we dive into each dataset and analyze trends locally ($\S$~\ref{sec:app_pred_perf_patterns_local}). 

\subsection{Aggregation Pipeline for Table~\ref{tab:relish_overall_main_with_statllm}}
\label{sec:app_agg_pipeline}

Table~\ref{tab:relish_overall_main_with_statllm} reports the macro-averaged results for Pearson correlation ($r$), Spearman correlation ($\rho$), and range-normalized Root Mean Squared Error (NRMSE). In this section, we specify the aggregation pipeline. Namely, we report macro-averaged means across metrics, combined with seed-to-seed standard deviations in $\pm$. We adopt this aggregation strategy to measure both the generalizability of regression-baselines across diverse tasks and LLMs and their stability across training runs.

\paragraph{Granular Metric Computation}
For each metric $\ell \in \{r, \rho, \text{NRMSE}\}$, we first compute the score $x_{d,m,s}^{(k, \ell)}$ for a given baseline $k$, evaluated on dataset $d$, using LLM backbone $m$, under some random seed $s$. Note that NRMSE normalizes RMSE by the label range of the respective dataset $d$ to ensure comparability across tasks with vastly different target scales:
$$\text{NRMSE}_{d,m,s}^{(k)} = \frac{\text{RMSE}_{d,m,s}^{(k)}}{y^{\max}_d - y^{\min}_d}$$
All reported metrics are scaled by $100$ to be reported as percentages.

\paragraph{Macro-averaging Across Datasets, LLM backbones, and Seeds}

For a given method $k$ and metric $\ell$, we compute a macro-mean $a_s^{(k,\ell)}$ for each seed $s$ by averaging the scores across all available dataset-model pairs $\mathcal{P}$:
$$a_s^{(k,\ell)} = \frac{1}{|\mathcal{P}|} \sum_{(d,m) \in \mathcal{P}} x_{d,m,s}^{(k, \ell)}$$

Table~\ref{tab:relish_overall_main_with_statllm} then reports the mean ($\mu$) and standard deviation ($\sigma$) of $a_s^{(k,\ell)}$:
$$\mu^{(k,\ell)} = \frac{1}{|\mathcal{S}|} \sum_{s \in \mathcal{S}} a_s^{(k,\ell)}$$
$$\sigma^{(k,\ell)} = \sqrt{\frac{1}{|\mathcal{S}|-1} \sum_{s \in \mathcal{S}} \left(a_s^{(k,\ell)} - \mu^{(k,\ell)}\right)^2}$$

\paragraph{Why not Micro-Average?}

We chose to report macro-averages rather than micro-averages, which would have first pooled individual predictions across all datasets before computing the respective metric. While this can be more intuitive, micro-averaging is highly susceptible to datasets with larger samples or higher variance, which can disproportionately skew the overall result. By computing the macro-mean $a_s^{(k,\ell)}$, we treat every dataset-model pair as an equal vote. Consequently, the superior predictive performance of {\method} in Table~\ref{tab:relish_overall_main_with_statllm} reflects robust and consistent gains across diverse datasets and LLM backbones, rather than potentially lifted by a few datasets or LLM backbones with large improvements.

In addition, we highlight that our reported $\sigma$ (seed-to-seed variance) is \textit{not} task heterogeneity, or the variance of metrics across different datasets and models. Task heterogeneity would be much larger, since datasets inherently vary in difficulty. Instead, our reported $\pm$ isolates stability across different seeds (training variance). The relatively tight $\pm$ bounds shown in Table~\ref{tab:relish_overall_main_with_statllm} confirm that the performance gains achieved by RELISH are statistically stable and not the artifact of a "lucky" random initialization.

\subsection{Additional Figures and Tables}
\label{sec:app_pred_perf_figures_and_tables}

\begin{figure}[ht]
    \centering
    \includegraphics[width=1\linewidth]{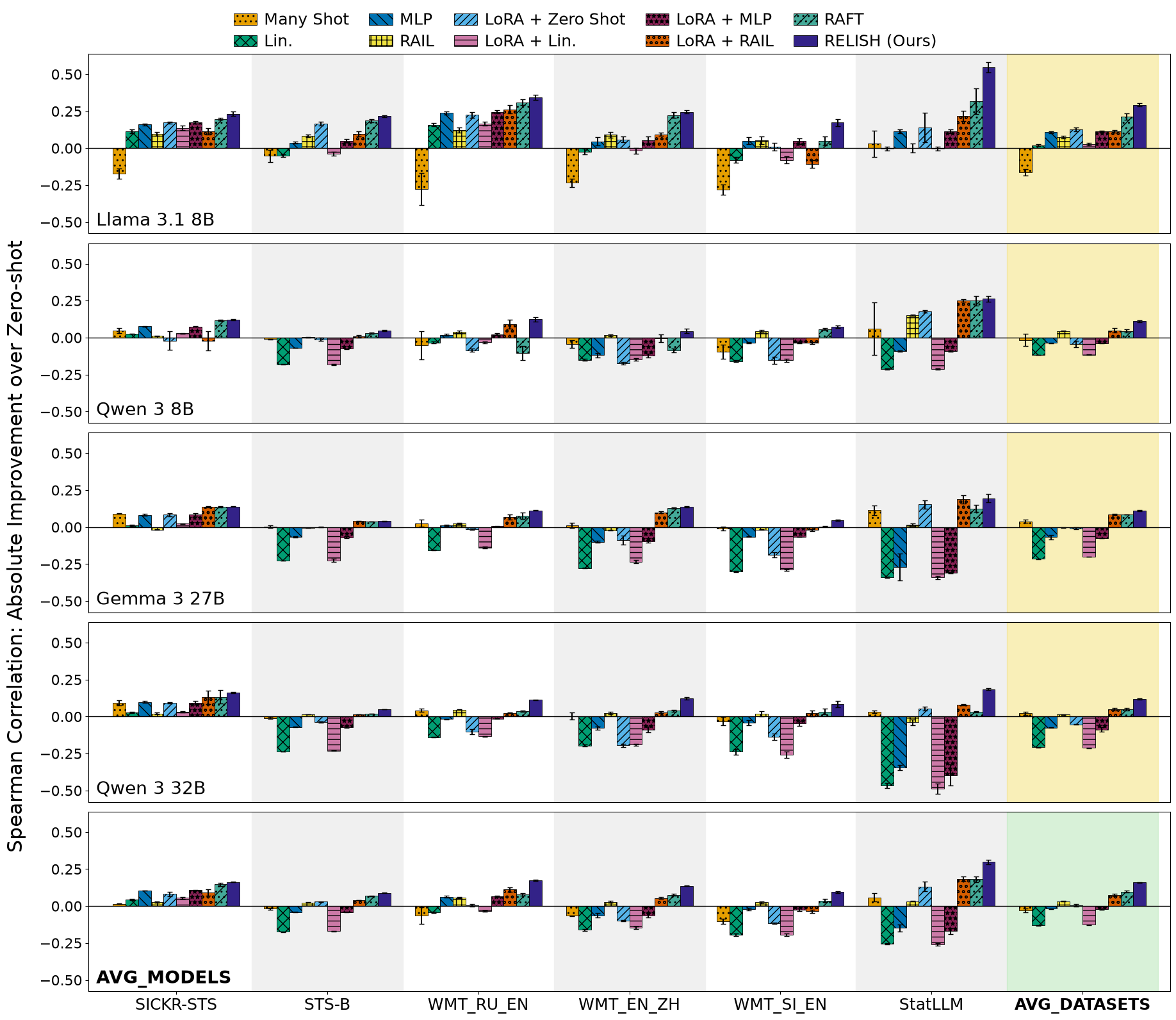}
\caption[\new{Spearman correlation across datasets and LLMs}]{\label{fig:spearman}\new{\textbf{Spearman ($\rho$) correlation across datasets (columns) and LLMs (rows).}} The rightmost column averages over all datasets, and the bottom row averages over LLMs. Whereas earlier Figure~\ref{fig:main_result_condensed} showed Pearson correlation, this figure shows absolute improvement in Spearman correlation over the zero-shot baseline (the "ground zero" line). Error bars denote variability across 3 independently-seeded runs. All metrics are computed using gold test labels. {\method} consistently performs best across all settings.}
\end{figure}

\begin{figure}[ht]
    \centering
    \includegraphics[width=1\linewidth]{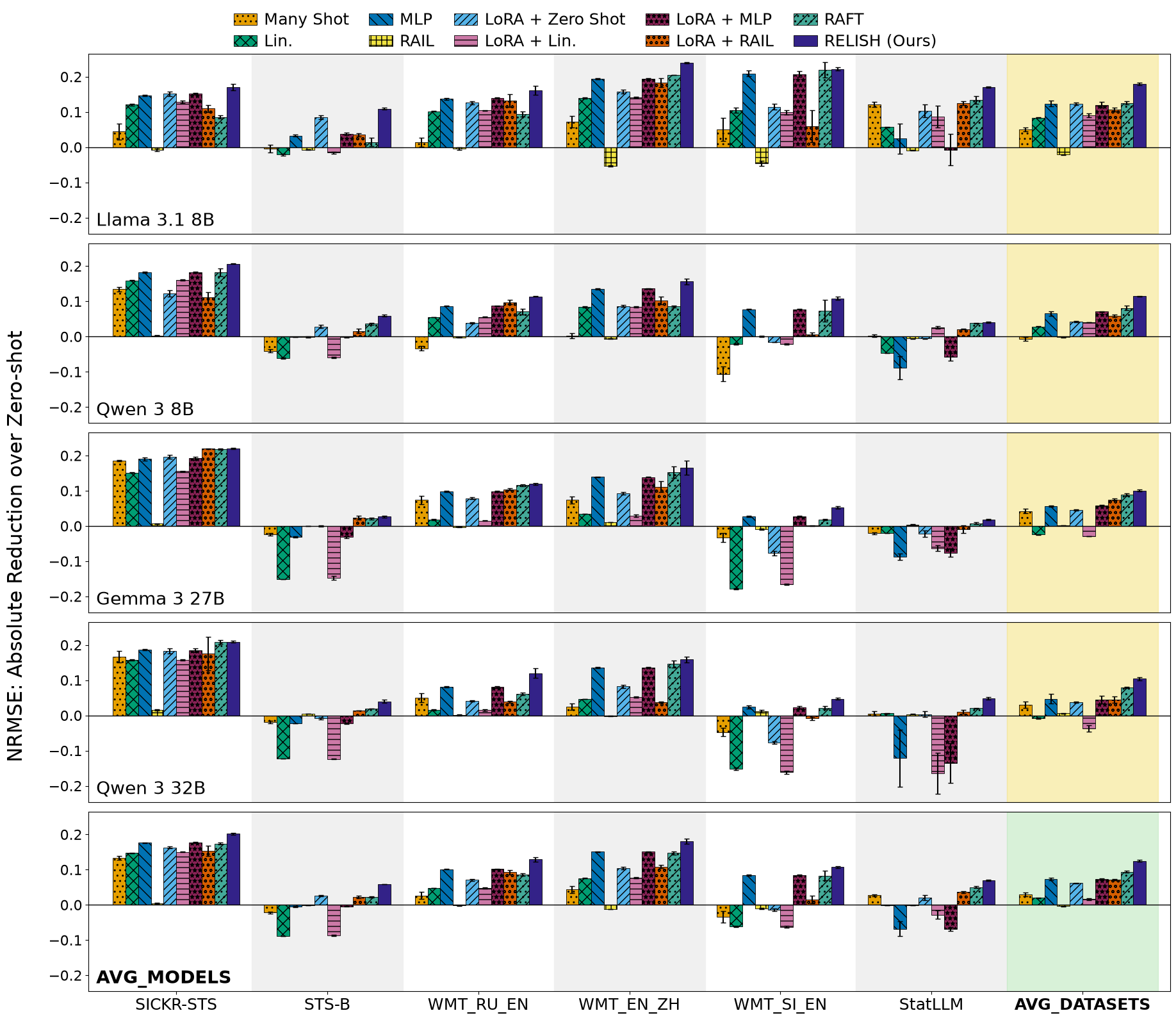}
\caption[\new{NRMSE results across datasets and LLMs}]{\label{fig:rmse}\new{\textbf{NRMSE results across datasets (columns) and LLMs (rows).}} The rightmost column averages over all datasets, and the bottom row averages over LLMs. Whereas earlier Figure~\ref{fig:main_result_condensed} showed Pearson correlation, this figure shows absolute improvement in range-normalized root mean squared error (NRMSE) over the zero-shot baseline (the "ground zero" line). Error bars denote variability across 3 independently-seeded runs. All metrics are computed using gold test labels. {\method} consistently performs best across all settings.}
\end{figure}

Figures~\ref{fig:spearman} and ~\ref{fig:rmse} report detailed results for Spearman correlation and the range-normalized root mean squared error (NRMSE) relative to the zero-shot performance of the underlying LLM backbones, respectively. For Figures~\ref{fig:main_result_condensed} and ~\ref{fig:spearman}, the absolute improvement is calculated as $r_{\method{}} - r_{zero\_shot}$ and $\rho_{\method{}} - \rho_{zero\_shot}$, respectively. For Figure~\ref{fig:rmse}, we flip the order and compute the absolute improvement as $NRMSE_{zero\_shot} - NRMSE_{\method{}}$, since lower is better.

Full predictive performance tables reporting raw RMSE, Pearson, and Spearman correlation across four LLM backbones and \new{six} datasets are reported in Tables~\ref{tab:stsb_relish_main}, \ref{tab:sickr_sts_relish_main}, \ref{tab:wmt_en_ru_relish_main}, \ref{tab:wmt_en_zh_relish_main}, \ref{tab:wmt_si_en_relish_main}, \new{and~\ref{tab:statllm_code_quality_relish_main}}. We do not report range-normalized RMSE there since each table compares baseline-LLM pairs within each dataset. Nevertheless, one may simply convert raw RMSE to NRMSE by dividing each dataset by its appropriate gold-score range (e.g., STS-B: 0–5, SICK-R: 1–5, WMT: 0–100\new{, StatLLM: 0--50}).

\subsection{High Level Patterns}
\label{sec:app_pred_perf_patterns_global}

\subsubsection{The Verbalization Gap: Why Prompting is Suboptimal}

Consistent with \citet{vacareanu2024from}, our results confirm that LLMs are capable regressors with learned numerical reasoning capabilities. Instead, the main challenge is how to \textit{best extract numerical predictions from LLMs}. Specifically, while pre-trained LLMs encode regression-relevant signals in their hidden states, existing inference methods are suboptimal at exploiting what the LLM already knows. 

Autoregressive (prompt-based) methods are the simplest, directly modeling numeric targets as text. Zero-shot prompting establishes a reasonable baseline (see Table~\ref{tab:relish_overall_main_with_statllm}), indicating that although LLMs are not explicitly trained as regressors, the internalized scale and relational priors (e.g., ``low" versus ``high") learned during pre-training are sufficient to verbalize approximate numeric values at inference time \citep{yuchi-du-eisner-2026}. 

However, autoregressive approaches are fundamentally limited by the \textbf{verbalization gap} \citep{yuchi-du-eisner-2026}: while LLM internal representations may accurately encode a scalar, the stochastic process of next-token generation often introduces noise, preventing the model from accurately translating its internal estimates into the correct string of digits. Our many-shot results (see Table~\ref{tab:relish_overall_main_with_statllm}) exemplify this gap. Contrary to expectations, adding in-context demonstration may actually degrade correlation metrics (e.g., $r=57.8$ compared to $r=61.0$ for zero-shot). We attribute this degradation to \textit{contextual distraction}: rather than calibrating the model's internal scale, the presence of fixed exemplars introduces formatting biases and a documented tendency to regress toward the mean of the provided demonstrations \citep{vacareanu2024from}. Consequently, simply providing more context fails to bridge the verbalization gap, as the model remains limited by the autoregressive generation process, which is poorly aligned with the numerical nature of regression targets. 

\subsubsection{The Summarization Bottleneck: Why Static Pooling for Predictive Heads is Suboptimal}

Predictive heads bypass the verbalization gap by treating the LLM as a feature extractor rather than a text generator. Specifically, predictive heads attach an auxiliary regressor to the model’s internal representations, aiming to directly "read out" numeric values that the autoregressive generator often fails to articulate. However, our results indicate that standard predictive heads are constrained by a \textbf{summarization bottleneck}: mean-pooled representations are often too coarse to capture the fine-grained signal necessary for regression.

For instance, our results show that a linear regressor over mean-pooled hidden states is the weakest baseline, significantly underperforming even zero-shot prompting in terms of Pearson and Spearman correlation (e.g., $r = 52.9$ versus $r = 61.0$, and $\rho = 50.3$ versus $\rho = 60.9$, respectively). We attribute this to \textit{information collapse}, where static averaging washes out the localized token-level features essential for precise numeric estimation (e.g., the log-magnitude signals identified by \citet{yuchi-du-eisner-2026}). While a non-linear MLP head ($r=64.7$, $\rho = 71.7$) can improve predictive performance, this approach remains bottlenecked by a fixed-vector summary and thus is suboptimal compared to {\method}. In summary, while existing predictive heads avoid the noise of token generation, they introduce their own limitations by compressing high-dimensional hidden states into static vector representations that are too restrictive for regression.

\subsubsection{Regression-aware Inference Helps, but Remains Suboptimal.}

Regression-aware baselines such as RAIL \citep{Lukasik2024-yt} and RAFT \citep{Lukasik2025-ak} attempt to bridge the verbalization gap by aligning the decoding process with continuous scalar predictions. These methods thus implicitly assume that LLMs already possess the internal reasoning capabilities for regression, but better verbalization techniques are needed to obtain accurate numerical estimates. However, regression-aware inference remains an indirect and computationally expensive solution to the extraction problem. 

Empirically, Table~\ref{tab:relish_overall_main_with_statllm} shows that RAIL can outperform zero-shot prompting on correlation metrics, but not on RMSE. A plausible explanation is that RAIL improves \emph{ranking} without improving \emph{scale calibration}. In particular, the constrained decoding procedure may preserve relative ordering across examples while shrinking predictions toward high-probability regions, clipping extreme values, or discretizing outputs too coarsely. Such effects can increase bias in the mean prediction even as they reduce variance, leading to better correlation but worse RMSE. More broadly, because RAIL still inherits the backbone LLM’s imperfect numeric calibration, averaging samples from its induced predictive distribution can stabilize outputs without correcting systematic errors in magnitude. We therefore view RAIL as a partial mitigation of the generation-regression mismatch, rather than a guaranteed improvement over zero-shot next-token prediction.

RAFT outperforms all autoregressive methods. However, its empirical gains are not consistent across all settings and can underperform baselines such as the MLP predictive head (e.g., see Figure~\ref{fig:main_result_condensed} for Qwen 3 8B on WMT\_RU\_EN). Computationally, RAFT requires an expensive pipeline: LoRA fine-tuning, regression-aware training, and regression-aware inference. 

In contrast, {\method} addresses the verbalization gap and summarization bottleneck more directly. Rather than forcing the generation process to behave like a regressor, {\method} learns a task-specific summarization over token-level representations to predict scores in a single pass. This approach is not only more computationally efficient but also more architecturally natural: if the log-magnitude signals are already present in the hidden states \citep{yuchi-du-eisner-2026}, directly learning to extract them is more intuitive than attempting to elicit them through sampled or constrained decoding.

\subsubsection{{\method} Provides Architectural Gains Beyond Head Capacity Alone.}

The gains from {\method} cannot be explained by simply adding more trainable parameters. Our MLP baseline is parameter-matched to {\method} for each LLM backbone, yet {\method} still consistently outperforms it across \new{six} datasets. This indicates that the improvements are architectural rather than capacity alone, suggesting that iterative token-level summarization is more effective for scalar prediction than static mean pooling. This interpretation is further supported by the comparison to the LoRA + zero-shot baseline. Although LoRA updates substantially more parameters, it remains less performant than {\method}. This suggests that, for regression tasks such as semantic sentence similarity and machine translation quality estimation, the primary bottleneck is not necessarily acquiring new task-relevant knowledge through backbone adaptation, but extracting an accurate scalar prediction from representations that already encode substantial numerical signal. From this perspective, LoRA may help with task adaptation by calibrating to expected output ranges, but it is less effective than {\method}'s iterative summarization mechanism.

\subsubsection{{\method} Provides Structural Gains Beyond Calibration Alone.}

The gains from {\method} are not solely due to calibration. Across Figures~\ref{fig:main_result_condensed}, \ref{fig:spearman}, and~\ref{fig:rmse}, {\method} consistently achieves the largest gains across Pearson, Spearman, and RMSE. This is significant because a calibration method that merely rescales predictions might lower RMSE or increase Pearson (e.g., LoRA fine-tuning), but such a method would not necessarily change the relative ranking of test queries. The consistent gains in Spearman correlation across LLM backbones and datasets demonstrate that {\method} actually improves the LLM's ability to rank instances by similarity or quality. Thus, {\method} is not just calibrating model predictions, but also learning how to make more accurate numerical predictions using a higher-resolution regression signal that is otherwise lost to the noise of verbalization or the coarseness of static pooling.

\subsubsection{\new{{\method} Helps Most When Regression Evidence is Localized or Distributed.}}

\new{Examining results at the task level helps clarify when {\method}'s architectural design provides the greatest benefit. The gains are most pronounced when regression-relevant evidence is localized or distributed across the input, rather than captured by a single global similarity signal. STS generally requires a global interpretation of semantic relatedness, which prompting and mean-pooled heads can often capture reasonably well. In contrast, WMT quality estimation depends on localized translation errors, omissions, and subtle meaning shifts that may be spread across source--translation pairs. StatLLM similarly requires identifying correctness, executability, and readability evidence across longer code inputs. Such fine-grained signals may be washed out by static pooling or obscured by autoregressive verbalization, whereas {\method} can repeatedly query token-level representations through cross-attention to preserve both localized and distributed cues.}

\subsection{Dataset Specific Patterns}
\label{sec:app_pred_perf_patterns_local}

\paragraph{STS-B (Table~\ref{tab:stsb_relish_main}):} For STS-B, improvements over baselines are generally modest. Because STS relies on global semantic similarity, larger LLM backbones already perform quite well with standard autoregressive or regression-aware decoding methods. Consequently, as the backbone scales, the headroom for improvement shrinks. This is evident on Gemma 3 27B Instruct, where {\method}'s Spearman correlation ($\rho=0.923$) effectively ties with the computationally heavier LoRA+RAIL baseline. Nevertheless, {\method} maintains a definite lead in both Pearson and RMSE. Thus, even where baseline methods are already strong, {\method} remains the most robust across all LLM backbones and metrics.

\paragraph{SICKR-STS (Table~\ref{tab:sickr_sts_relish_main}):} Similarly, for SICKR-STS, improvements over baselines are generally modest. However, in this dataset, predictive heads are surprisingly strong baselines, suggesting that the mean-pooled representations retained much of the information embedded in full token-level hidden states. Yet, even in this highly competitive setting, {\method} consistently squeezes additional performance across all four backbones. For instance, on Gemma 3 27B Instruct, {\method} narrowly but definitively outperforms the strongest alternatives, including RAFT and LoRA+RAIL. This demonstrates that even when existing baselines perform reasonably well, iterative latent refinement still captures valuable, task-relevant signals that are missed by verbalization and static aggregation.

\paragraph{WMT Ru–En (Table~\ref{tab:wmt_en_ru_relish_main}):} RELISH’s architectural advantages are especially evident on WMT\_RU\_EN. This dataset exposes the weaknesses of generation-based methods: many-shot prompting yields inconsistent gains, and RAIL frequently trails behind learned predictive heads. RELISH capitalizes on this, achieving substantial margins over RAFT and LoRA baselines across all backbones. This trend reinforces our prior hypothesis that machine translation quality estimation requires selectively analyzing localized cues spread across source-translation pairs, a task {\method} excels at with its token-level latent querying.

\paragraph{WMT En–Zh (Table~\ref{tab:wmt_en_zh_relish_main}):} This translation dataset poses the most difficult task in our suite, yielding significantly lower correlation metrics across all methods and LLM backbones. Under these challenging conditions, standard methods may not perform well. Notably, even LoRA-adapted methods (particularly LoRA+Zero Shot and LoRA+Linear) degrade sharply, struggling to maintain stability. RAFT, typically the second-best baseline, may actually underperform simpler baselines like LoRA+RAIL on Qwen 3 8B. In contrast, {\method} remains highly robust, consistently achieving the highest predictive performance. 

\paragraph{WMT Si-En (Table~\ref{tab:wmt_si_en_relish_main}):} WMT\_SI\_EN is meant to test the capabilities of LLMs on a low-resource language setting, such as translating from Sinhala to English. Interestingly, the correlation scores here are higher than those in WMT\_EN\_ZH. For instance, {\method} reaches the highest Pearson correlation of $r=0.669$ on Gemma 3 27B Instruct here, versus a best of $r=0.567$ on Qwen 3 32B for WMT\_EN\_ZH. We suspect that since the translation is in English, the LLM backbones can rely on strong English priors to avoid major structural errors that often appear in low-resource translation \citep{guzman2019flores, han2021translation}. Nevertheless, {\method} performs best across all four backbones, suggesting that its token-level summarization remains effective even under low-resource settings.

\paragraph{\new{StatLLM (Table~\ref{tab:statllm_code_quality_relish_main}):}} As noted in 
$\S$~\ref{sec:datasets_setup_extended}, \new{StatLLM provides a qualitatively different test case because code generation is a more complex task, and because each example consists of a generated SAS program plus task context, rather than a sentence pair or source--translation pair. That said, we have also noted the smaller scale of this dataset compared to others, and future work on larger scale code generation datasets may reveal additional patterns.}

\new{
{\method} is seen to achieve the best 
results across all four LLM backbones. This suggests that {\method}'s gains are not solely due to cross-attention aligning two input segments. Instead, the results support our broader interpretation that {\method} improves regression by learning which parts of the input are most informative, including localized correctness issues and distributed readability signals in longer single-input programs.}

\section{Full Training Footprint Results}
\label{sec:app_footprint}

Table~\ref{tab:lightweight_detailed} compares the exact trainable component count across all evaluated LLM backbones for {\method}, LoRA \citep{hu2022lora}, and RAFT \citep{Lukasik2025-ak}. Notably, while Table~\ref{tab:lightweight} and Table~\ref{tab:lightweight_detailed} show that LoRA and RAFT have the same trainable parameter count, in practice, RAFT incurs a higher cost because the posterior for each training example must still be approximated via sampling or enumerating grid candidates. 

Despite being the most performant method, {\method} also fine-tunes significantly fewer parameters than LoRA and RAFT, which use a regression-aware loss for LoRA fine-tuning. For instance, on the 8B-class models (Llama 3.1 and Qwen 3), {\method} introduces merely 3.4 million trainable parameters, or roughly 0.04\% of the LLM backbone. In contrast, standard LoRA and RAFT adaptation requires over 20 million parameters, or roughly 0.26\% of the LLM backbone, representing a roughly six-fold increase in parameter overhead. This gap is even more significant for larger models. For Qwen 3 32B, LoRA and RAFT require fine-tuning roughly 37 times as many parameters as {\method}. 

In addition, unlike LoRA and RAFT, {\method}'s training footprint does \textbf{not} scale with the LLM backbone. Rather, the dominant components in {\method} are the projection matrix $W \in \mathbb{R}^{d \times d_h}$ and the iterative refinement layers consisting of the multi-head attention (MHA), the FFN, and the LayerNormalization (LN). Thus, the main hyperparameters that affect the trainable capacity of {\method} are the head dimension $d_h$, the number of refinement layers $L$, and parameter sizes within each refinement layer, such as the hidden size of the FFN ($4 * d_h$). Conversely, this means that {\method}'s training footprint \textbf{does not scale with the size of the LLM backbone}. In summary, {\method} is not performant merely because it fine-tunes more parameters, but by providing a more architecturally efficient summarization mechanism.

\begin{table*}[ht]
\centering
\small
\setlength{\tabcolsep}{12pt} 
\renewcommand{\arraystretch}{1.2}
\begin{tabular}{l r c}
\toprule
\textbf{Method} & \textbf{Trainable ($n$)} & \textbf{\% of Model} \\
\midrule

\multicolumn{3}{l}{\cellcolor{gray!10}\textbf{Llama 3.1 8B Instruct} \quad ($d=4096, N=8,030,261,248$)} \\
LoRA/RAFT & 20,971,520 & 0.2612\% \\
\textbf{{\method}} & \textbf{3,418,625} & \textbf{0.0426\%} \\
\addlinespace[0.6em]

\multicolumn{3}{l}{\cellcolor{gray!10}\textbf{Qwen 3 8B} \quad ($d=4096, N=8,190,735,360$)} \\
LoRA/RAFT & 21,823,488 & 0.2664\% \\
\textbf{{\method}} & \textbf{3,418,625} & \textbf{0.0417\%} \\
\addlinespace[0.6em]

\multicolumn{3}{l}{\cellcolor{gray!10}\textbf{Gemma 3 27B Instruct} \quad ($d=5376, N=27,432,406,640$)} \\
LoRA/RAFT & 116,502,528 & 0.4247\% \\
\textbf{{\method}} & \textbf{3,746,305} & \textbf{0.0137\%} \\
\addlinespace[0.6em]

\multicolumn{3}{l}{\cellcolor{gray!10}\textbf{Qwen 3 32B} \quad ($d=5120, N=32,762,123,264$)} \\
LoRA/RAFT & 134,217,728 & 0.4097\% \\
\textbf{{\method}} & \textbf{3,680,769} & \textbf{0.0112\%} \\
\bottomrule
\end{tabular}
\caption[Detailed parameter footprints across architectures]{\textbf{Detailed parameter footprints across architectures.} $N$ represents the total number of backbone parameters and $d$ is the model hidden dimension. \textbf{Trainable ($n$)} is the number of parameters updated during training, and \textbf{\% of Model} denotes $n/N$. Note that while baseline footprints scale with the total parameter count $N$, {\method}'s footprint remains largely invariant to backbone size, scaling primarily as a function of $d$.}
\label{tab:lightweight_detailed}
\end{table*}

\section{Ablation Results}
\label{sec:app_ablations}

\subsection[How sensitive is RELISH to the choice of loss function?]{How sensitive is {\method} to the choice of loss function?}
\label{sec:app_ablations_loss}

\begin{table}[ht]
\centering
\small
\setlength{\tabcolsep}{2.0pt}
\renewcommand{\arraystretch}{1.25}

\begin{tabular*}{\columnwidth}{@{\extracolsep{\fill}} l cc cc @{}}
\toprule
& \multicolumn{2}{c}{\textbf{Fine-tuned Baselines}} & \multicolumn{2}{c}{\textbf{RELISH Variants}} \\
\cmidrule{2-3} \cmidrule{4-5}
\textbf{Metric} & \textbf{LoRA+MLP} & \textbf{RAFT} & \textbf{RELISH (MSE)} & \textbf{RELISH (Huber)} \\
\midrule
\textbf{$r$} $\uparrow$ & $64.8_{\scriptscriptstyle \pm 0.1}$ & $71.5_{\scriptscriptstyle \pm 0.4}$ & $74.2_{\scriptscriptstyle \pm 0.4}$ & \cellcolor{green!20} $\mathbf{76.3_{\scriptscriptstyle \pm 0.2}}$ \\
\textbf{$\rho$} $\uparrow$ & $61.7_{\scriptscriptstyle \pm 0.1}$ & $69.0_{\scriptscriptstyle \pm 0.2}$ & $71.3_{\scriptscriptstyle \pm 0.4}$ & \cellcolor{green!20} $\mathbf{74.0_{\scriptscriptstyle \pm 0.1}}$ \\
\textbf{NRMSE} $\downarrow$ & $16.7_{\scriptscriptstyle \pm 0.1}$ & $16.6_{\scriptscriptstyle \pm 0.3}$ & $13.5_{\scriptscriptstyle \pm 0.1}$ & \cellcolor{green!20} $\mathbf{13.3_{\scriptscriptstyle \pm 0.3}}$ \\
\bottomrule
\end{tabular*}
\caption[Ablating RELISH with Huber versus MSE loss]{Ablating {\method} with Huber versus MSE loss. We report macro-averaged Pearson $r$ and Spearman $\rho$ correlation (higher is better), as well as range-normalized RMSE (NRMSE, lower is better) in \%. Metrics are averaged across \new{the five STS/WMT datasets, excluding StatLLM}, four LLM backbones, and three independently seeded runs. We also report standard deviation ($\pm$).}
\label{tab:ablation_loss}
\end{table}

\begin{figure}
    \centering
    \includegraphics[width=1.0\linewidth]{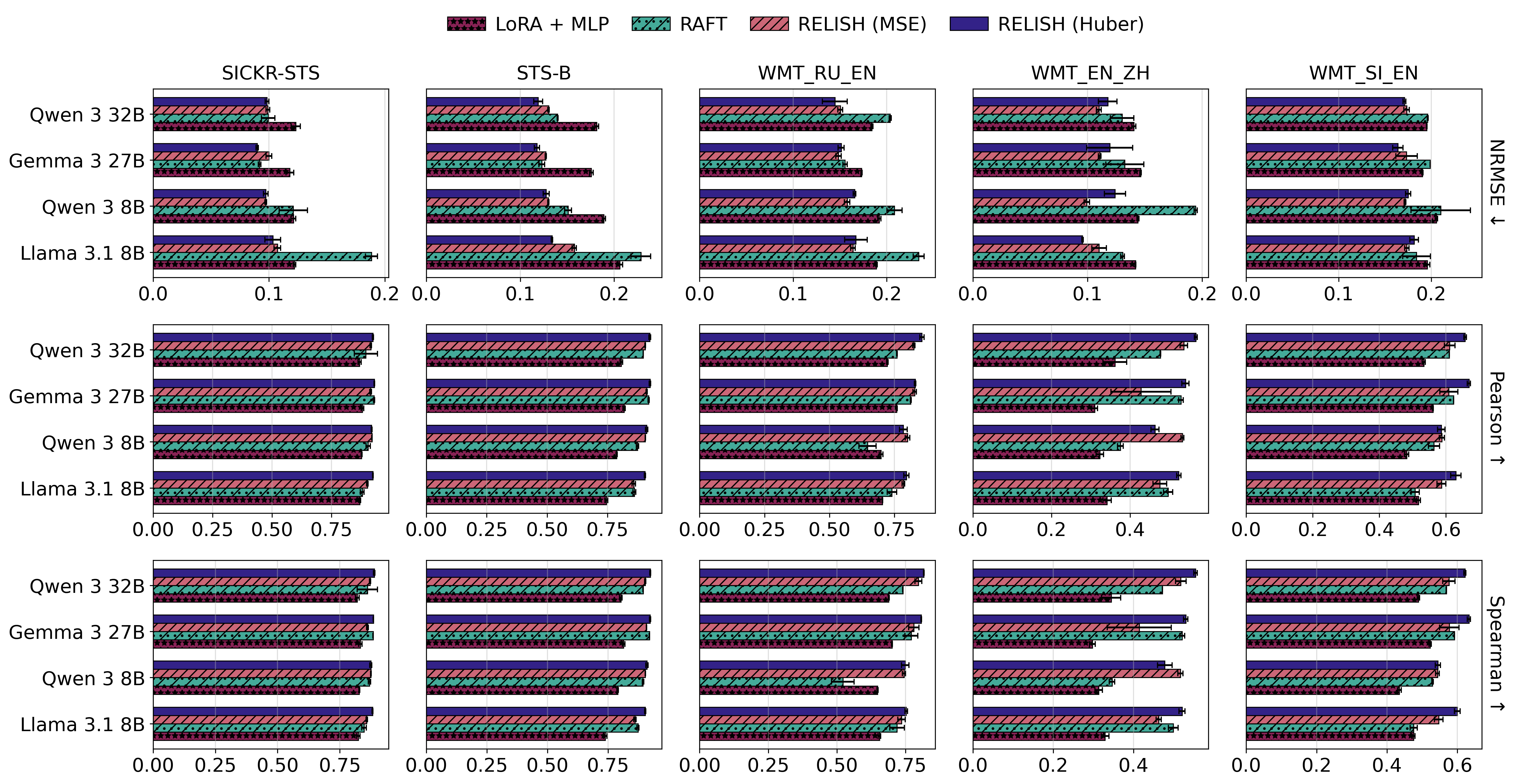}
    \caption[RELISH loss ablations]{\textbf{{\method} loss ablations}. We compare LoRA + MLP, RAFT, and two {\method} variants trained with MSE or Huber across \new{the five STS/WMT datasets, excluding StatLLM} (columns) and four LLM backbones (rows). Metrics are NRMSE (lower is better), Pearson correlation (higher is better), and Spearman correlation (higher is better). Error bars show variability across 3 seeds. Overall, MSE and Huber perform similarly, but Huber is modestly better in most settings.}
    \label{fig:ablation_loss}
\end{figure}

Table~\ref{tab:relish_overall_main_with_statllm} shows that {\method} consistently improves over prior LLM-regression baselines. But where do the gains come from? In this section, we investigate \textbf{whether the improvements are due to better architectural design or simply an artifact of the Huber loss} \citep{huber1992robust}. \new{We conduct ablations on the five STS/WMT datasets, excluding StatLLM.} \new{Another motivation is to enable fairer comparisons with RAFT \citep{Lukasik2025-ak}. While RAFT currently employs a squared loss (Table~\ref{tab:regression-taxonomy}), RAFT could potentially use a Huber formulation by penalizing large prediction errors with a L1 loss.}

Thus, we ablate {\method} trained with the standard mean-squared error (MSE) loss. We keep y-normalization, as we observe that training collapses without it. Specifically, without label normalization, all predictive heads will default to predicting the dataset label mean as the simplest solution to minimize numerical error. 

Theoretically, the Huber loss helps stabilize training by behaving like MSE near the numeric target while down-weighting occasional large residuals, which is attractive in our setting because our evaluation spans heterogeneous tasks with different score ranges and annotation processes. At the same time, since label normalization already caps large residuals, we hypothesize that the performance difference between {\method} trained with Huber and with MSE will be modest. 

Following the aggregation protocol detailed in $\S$~\ref{sec:app_agg_pipeline}, we report macro-averaged metrics and standard deviations in Table~\ref{tab:ablation_loss}. For reference, we duplicate the two most performant baselines from Table~\ref{tab:relish_overall_main_with_statllm}, namely LoRA + MLP and RAFT \citep{Lukasik2025-ak}. We then provide more fine-grained result breakdowns per dataset and LLM backbone in Figure~\ref{fig:ablation_loss}.

Based on Table~\ref{tab:ablation_loss}, on average, we observe that \textbf{the improvements from {\method} are likely due to improved architectural design, and not an artifact of the Huber loss}. Compared to the other baselines, both variants of {\method} yield higher overall predictive performance. Notably, on NRMSE, {\method} trained with Huber or MSE is very similar (13.5 vs. 13.3), but the Huber loss yields greater improvements on the correlation metrics (76.3 vs. 74.2, 74.0 vs. 71.3, respectively). Nevertheless, the main gains are likely from the iterative-refinement design of {\method}, and using Huber over MSE yields modest improvements in addition to the gains from the architecture. 

Figure~\ref{fig:ablation_loss} confirms the overall pattern observed in Table~\ref{tab:ablation_loss}. Across \new{the five STS/WMT datasets}, four LLM backbones, and three metrics, the {\method} (MSE) and {\method} (Huber) bars are usually close, often nearly overlapping, indicating that \textbf{{\method} is generally robust to the loss function}. This affirms our prior intuition that after label normalization, extreme residual magnitudes become less dominant, so Huber no longer has a structural advantage over MSE. Nevertheless, for the majority of panels, {\method} (Huber) outperforms {\method} (MSE), slightly scoring lower for NRMSE and higher for Pearson and Spearman. This suggests that although Huber does not help with output calibration, since labels are already normalized, it can help {\method} learn cleaner monotonic relationships when ranking test queries. 

Looking more closely at Figure~\ref{fig:ablation_loss}, there are also settings where {\method} with MSE outperforms the original Huber variant. Most notably, for Llama 3.1 8B and Qwen 3 8B, {\method} (MSE) ties with or exceeds {\method} (Huber) across several translation datasets. Similar reversals occasionally appear for larger LLM backbones, but less frequently. We rarely observe any reversals for the semantic similarity datasets. 

One possible explanation is that smaller LLM backbones are more capacity-limited, so their main bottleneck is representational expressivity rather than robustness to occasional large residuals. Hence, the fully quadratic signal from MSE may sometimes be more preferable, especially on noisier translation tasks where fitting fine-grained score variation is already difficult. In contrast, larger LLM backbones may be better able to exploit Huber’s robustness. Once the representation is sufficiently strong, down-weighting harder or noisier examples can more reliably improve ranking consistency. Of course, the main empirical trend remains unchanged: {\method} performs strongly under both losses, and Huber is better overall and in most settings, but only modestly so.

Finally, we note that \textbf{{\method}'s gains are not due to using a more robust objective than other regression baselines}. On the contrary, the baselines we compare to in Table~\ref{tab:relish_overall_main_with_statllm} typically employ loss functions that are at least as robust as Huber. For the predictive head methods (linear and MLP), we use the same Huber objective. For RAFT, a special regression-aware loss is employed. The only exception might be the LoRA fine-tuned autoregressive methods, but their capacity is larger (more trainable parameters), and yet {\method} consistently outperforms them all. Hence, the gains from {\method} cannot be explained solely by the Huber loss. The iterative architecture offers a clear advantage over existing LLM-regression methods.  

\subsection[How sensitive is RELISH to the number of iterative refinement blocks?]{How sensitive is {\method} to the number of iterative refinement blocks (\texorpdfstring{$L$}{L})?}
\label{sec:app_ablations_L}

\begin{figure}
    \centering
    \includegraphics[width=1.0\linewidth]{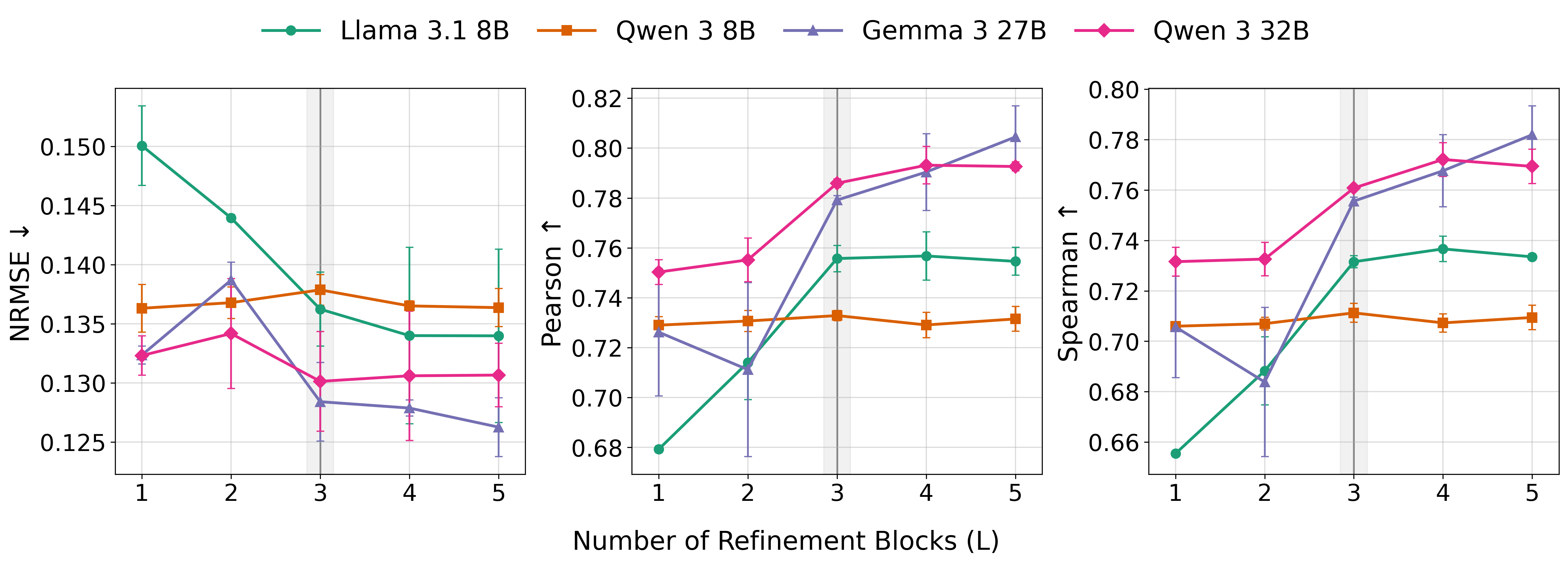}
    \caption[RELISH refinement depth ablations]{\textbf{{\method} refinement depth ablations.} We systematically vary the refinement depth of {\method} from 1 to 5. We report range-normalized root mean square error (NRMSE; lower is better), Pearson correlation, and Spearman correlation (higher is better) macro-averaged across four LLM backbones and \new{the five STS/WMT datasets, excluding StatLLM}. Error bars denote the standard deviation across three random seeds. The $L=1$ configuration isolates the performance of a single-step, attention-based pooling baseline. For most LLM backbones, multiple refinement iterations ($L \ge 2$) substantially enhance predictive performance, with diminishing returns past $L=3$ (our default setting, shaded in grey). Notably, the optimal depth varies by LLM architecture: Qwen 3 8B achieves peak performance at $L=1$, while the larger Gemma 3 27B continues to benefit from refinement even at higher depths.}
    \label{fig:ablations_L}
\end{figure}

Previously, $\S$~\ref{sec:app_ablations_loss} demonstrated that the improvements from {\method} are primarily driven by architectural design rather than the Huber loss, which only modestly improves predictive performance. Furthermore, $\S$~\ref{sec:relish_diff} highlighted two core architectural differentiations of {\method} from prior predictive head methods: learning a latent attention-based query for robust summarization, and iteratively refining this latent state. In this section, we aim to \textbf{disentangle the observed gains of {\method} between the attention-pooling mechanism and the iterative nature of the refinement process}. 

Specifically, we systematically vary the number of refinement blocks, $L$, from 1 to 5. Notably, the $L=1$ configuration serves as a strict learned, attention-based query-pooling baseline, as the output prediction is computed in a single pass without subsequent iterations. By comparing $L=1$ against variants where $L \ge 2$, we can explicitly isolate the performance gains attributable to repeated latent state updates versus attention-pooling alone. 

Figure~\ref{fig:ablations_L} reports LLM-specific metrics macro-averaged across \new{the five STS/WMT datasets, excluding StatLLM}, with $\pm$ denoting standard deviation across seeds. The primary takeaway is that \textbf{although attention-pooling alone serves as a strong baseline, iterative refinement blocks substantially enhance predictive performance}. While a single pass of token-level cross-attention ($L=1$) outperforms naive pooling, it remains suboptimal compared to the full {\method} architecture. Across the evaluated LLM backbones and datasets, the $L=1$ configuration generally yields the highest range-normalized root mean square error (NRMSE; lower is better) and the lowest Pearson and Spearman correlations (higher is better). This confirms that merely attending to token representations is insufficient to fully capture the regression target from the frozen latent space. As $L$ increases, we observe sharp reductions in NRMSE alongside notable improvements in both correlation metrics, demonstrating that the latent state requires at least one subsequent pass to re-attend to the token sequence and properly contextualize its initial estimate.

Following the initial performance surge at $L=2$, we observe a clear pattern of convergence and diminishing returns. The NRMSE curves rapidly flatten, and all predictive metrics become highly stable across $L=3$ through $L=5$, exhibiting only marginal fluctuations. Collectively, these patterns justify our default choice of $L=3$. Importantly, {\method} does not exhibit optimization instabilities or degradation at these deeper refinement levels.

However, we note distinct exceptions to these trends across model architectures. For Qwen 3 8B, iterations yield minimal benefits; the model achieves its lowest NRMSE at $L=1$, with performance slightly degrading as refinement blocks are added. This suggests that Qwen 3 8B's frozen representations may already encode regression features in a highly localized, accessible manner, allowing efficient extraction in a single pass, whereas subsequent iterations merely inject noise or induce overfitting. We do not observe this behavior in the larger Qwen 3 32B model. We hypothesize this discrepancy stems from the 32B model's larger hidden dimension (5120 vs. 4096), which may dilute regression-relevant information and necessitate iterative contextualization. 

Conversely, while Gemma 3 27B Instruct clearly benefits from added iterations, its predictive performance does not plateau at higher depths. Instead, all three metrics continue to improve through $L=5$, suggesting further gains are possible. We suspect this is similarly tied to the model's hidden size; Gemma 3 27B features the largest hidden dimension in our suite (5376; see Table~\ref{tab:lightweight_detailed}), necessitating deeper iterative refinement to fully isolate task-relevant signals. Therefore, while $L=3$ serves as a robust empirical default, the optimal refinement depth is ultimately intrinsically tied to the latent representation geometry of the specific LLM backbone.

\subsection[How much supervised data does RELISH need?]{\new{How much supervised data does {\method} need?}}
\label{sec:app_ablations_data}

\new{
Since {\method} trains parameters from scratch, a natural consideration is \textbf{how much supervised data {\method} requires}. Towards this end, we now investigate how regression performance scales with supervised data availability on the STS-B and WMT\_EN\_ZH datasets. We subsample the training set at 64, 128, 256, 512, 1,024, and 2,048 training instances, while keeping the validation and test sets fixed. We also replicate the full-dataset performance reported in our paper for reference (5749 for STS-B and 7000 for WMT\_EN\_ZH). We conduct experiments across the same four LLMs and three random seeds as detailed in $\S$~\ref{sec:experimental_setup_extended}.
}

\new{
We compare {\method} against MLP and RAFT \citep{Lukasik2025-ak}. MLP is the strongest predictive head baseline, while RAFT is the strongest supervised regression-aware baseline. For a method-level comparison, we report the RMSE, Pearson, and Spearman correlation averaged across four LLM backbones and three random seeds. To ensure fairness, sampled subsets are nested so that trials with more training instances always include those with fewer (e.g., the 128-example set must contain the 64-example set). 
}

\begin{figure*}[t]
    \centering
    \includegraphics[width=\textwidth]{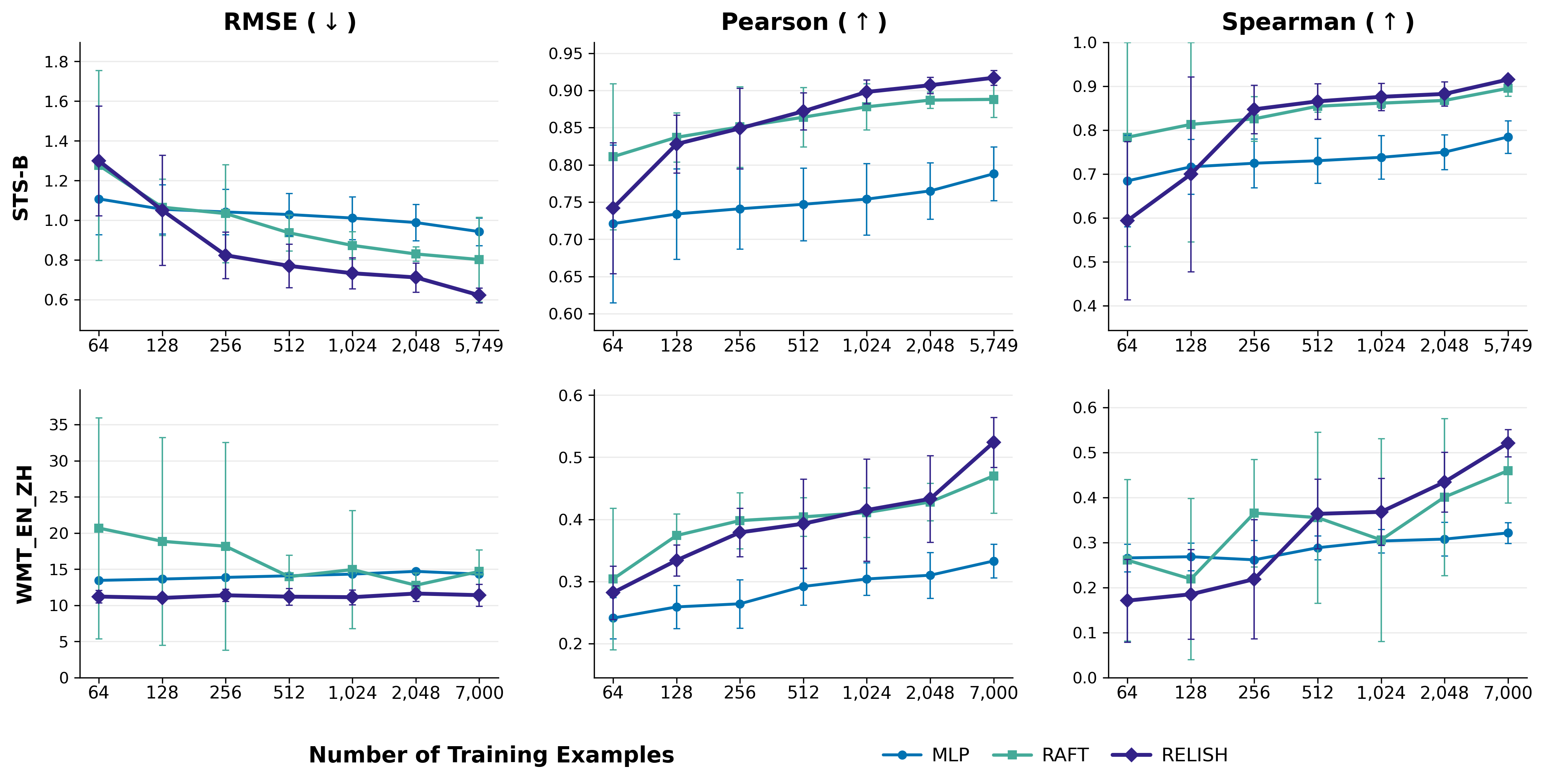}
    \caption[How predictive performance scales with supervised data availability.]{\new{\textbf{How predictive performance scales with data availability.} We compare {\method} against MLP and RAFT on STS-B and WMT\_EN\_ZH as the number of training examples increases from 64 to the full training set (5,749 for STS-B and 7,000 for WMT\_EN\_ZH). We report RMSE (lower is better), Pearson correlation (higher is better), and Spearman correlation (higher is better), averaged across four LLM backbones and three random seeds. Error bars denote standard error across the corresponding 12 model-seed runs. RAFT performs best under extremely low data regimes ($<$ 256 examples), but {\method} benefits more from additional supervision ($>$ 256--512 examples).}}
    \label{fig:ablation_data_perf}
\end{figure*}

\new{
Figure~\ref{fig:ablation_data_perf} shows that predictive performance for each method varies depending on the metric and the amount of labeled data. On correlation metrics, MLP consistently performs the worst in all cases, indicating that mean-pooled predictive heads struggle to rank examples accurately even with more supervision. RAFT performs best in scarce data regimes: on both STS-B and WMT\_EN\_ZH, RAFT achieves the best Pearson and Spearman correlations with 64--256 examples. However, {\method} improves more steadily with additional supervision. On STS-B, {\method} surpasses RAFT by 512 examples and continues to widen the gap as the number of training instances increases. On WMT\_EN\_ZH, {\method} remains competitive up to 512 examples, but outperforms RAFT with 1,024 or more examples.} 

\new{
The RMSE trends are slightly different. {\method} achieves the lowest RMSE in most cases, except for scarce data settings on STS-B (64 and 128 examples). Overall, RAFT is more performant when labels are extremely scarce ($<$ 256 examples), whereas {\method} benefits more from moderate-to-large ($>$ 256--512 examples) supervised datasets. Thus, a promising future direction might be to explore whether further improvements are possible by combining the best aspects of RELISH and RAFT (see $\S$~\ref{sec:app_limitations}).
}

\subsection[How does training overhead scale with supervised data?]{\new{How does training overhead scale with supervised data?}}
\label{sec:app_ablations_runtime}

\begin{figure*}[t]
    \centering
    \includegraphics[width=0.95\textwidth]{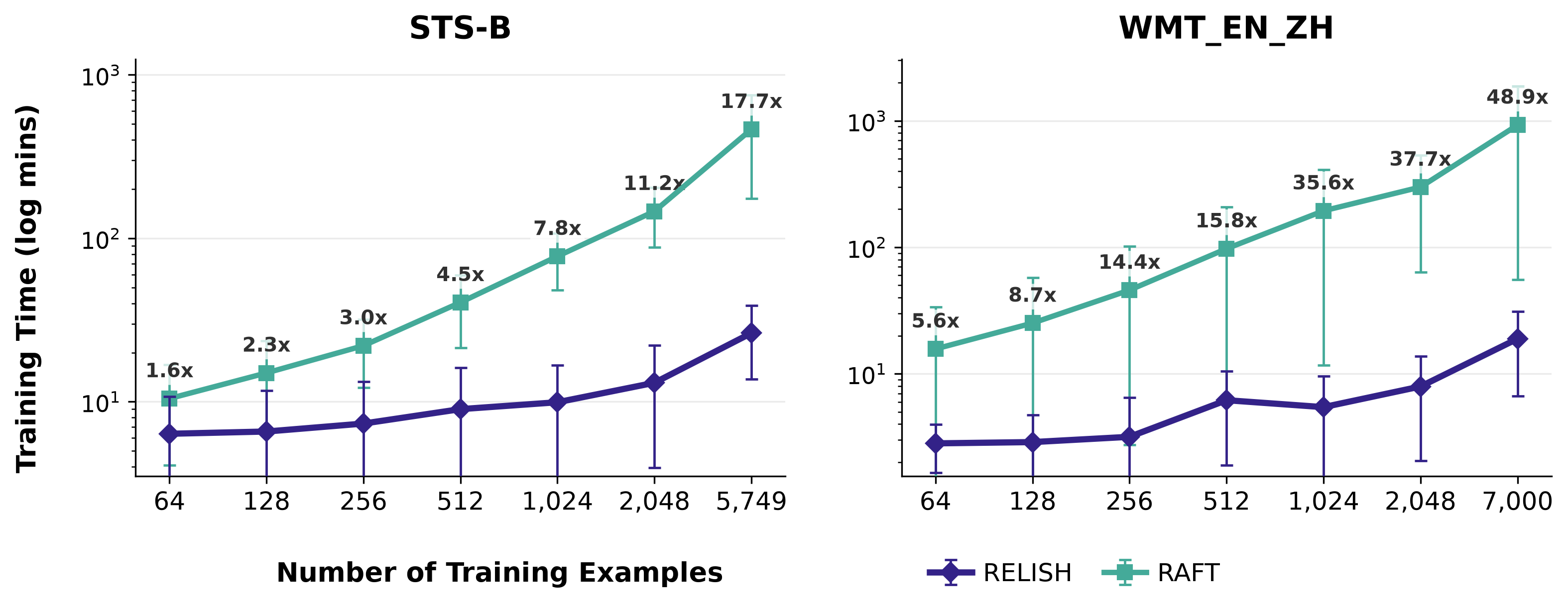}
    \caption[Training overhead across different data scales]{\new{\textbf{Training overhead across different data scales.} We compare {\method} and RAFT on STS-B and WMT\_EN\_ZH as the number of training examples increases from 64 to the full training set. Training time is reported in minutes on a log scale, averaged across four LLM backbones and three random seeds. Error bars denote standard error across 12 model-seed runs, and text labels report the RAFT/{\method} runtime ratio at each training size. {\method} is consistently faster to train than RAFT.}}
    \label{fig:ablation_training_runtime}
\end{figure*}

\new{
We also measure the training overhead of {\method} versus RAFT across different supervision regimes on STS-B and WMT\_EN\_ZH, following the setup specified in $\S$~\ref{sec:app_ablations_data}.
Specifically, we benchmark the total training wall-clock on a compute node with an Intel Xeon Platinum 8468 (``Sapphire Rapids") CPU and 4x NVIDIA H100.
Figure~\ref{fig:ablation_training_runtime} shows that RAFT is consistently slower to train than {\method}, and the gap widens as the number of training instances grows. 
Even with only 64 examples, RAFT is already $1.6\times$ slower on STS-B and $5.6\times$ slower on WMT\_EN\_ZH. Using the full training dataset split, the gap grows to $17.7\times$ on STS-B and $48.9\times$ on WMT\_EN\_ZH. 
Together with Figure~\ref{fig:ablation_data_perf}, these results suggest that while RAFT performs best when labels are extremely scarce, {\method} might still be preferred for its lightweight architecture. Once moderate supervision is available, {\method} is both more accurate and more efficient to train.
} 

\subsection[How efficient is RELISH at inference time?]{\new{How efficient is {\method} at inference time?}}
\label{sec:app_ablations_inference_time}

\new{
Related to training efficiency, we also investigate \textbf{how efficient {\method} is at inference time}. {\method} makes one forward pass through the LLM backbone and emits a prediction using a lightweight predictive head, but the additional refinement layers may impose additional overhead compared to a simple MLP head. Toward this end, we measure inference wall-clocks for all benchmarked LLM-based regression baselines on the test sets of STS-B and WMT\_EN\_ZH. All experiments are conducted on a compute node with an Intel Xeon Platinum 8468 (``Sapphire Rapids") CPU and 4x NVIDIA H100. 
}

\begin{table*}[t]
\centering
\small
\setlength{\tabcolsep}{9pt}
\renewcommand{\arraystretch}{1.15}
\begin{tabular}{llcc}
\toprule
\textbf{Dataset} & \textbf{Method} & \textbf{Throughput (ex/s)} $\uparrow$ & \textbf{Latency/{\method}} $\downarrow$ \\
\midrule
\multirow{7}{*}{STS-B} & MLP & $15.58_{\scriptscriptstyle \pm 8.63}$ & $0.99\times$ \\
                       & Linear & $15.57_{\scriptscriptstyle \pm 8.61}$ & $0.99\times$ \\
                       & \textbf{{\method}} & $15.35_{\scriptscriptstyle \pm 8.36}$ & $1.00\times$ \\
                       & Zero-shot & $3.17_{\scriptscriptstyle \pm 2.53}$ & $4.85\times$ \\
                       & RAFT & $2.92_{\scriptscriptstyle \pm 1.70}$ & $5.26\times$ \\
                       & RAIL & $1.24_{\scriptscriptstyle \pm 0.78}$ & $12.43\times$ \\
                       & Many-shot & $1.03_{\scriptscriptstyle \pm 0.71}$ & $14.95\times$ \\
\cmidrule(lr){1-4}
\multirow{7}{*}{WMT\_EN\_ZH} & MLP & $15.57_{\scriptscriptstyle \pm 8.53}$ & $0.99\times$ \\
                           & Linear & $15.57_{\scriptscriptstyle \pm 8.57}$ & $0.99\times$ \\
                           & \textbf{{\method}} & $15.34_{\scriptscriptstyle \pm 8.29}$ & $1.00\times$ \\
                           & Zero-shot & $2.38_{\scriptscriptstyle \pm 1.97}$ & $6.45\times$ \\
                           & RAFT & $1.17_{\scriptscriptstyle \pm 0.98}$ & $13.12\times$ \\
                           & RAIL & $1.08_{\scriptscriptstyle \pm 0.69}$ & $14.21\times$ \\
                           & Many-shot & $0.62_{\scriptscriptstyle \pm 0.44}$ & $24.86\times$ \\
\bottomrule
\end{tabular}
\caption[\new{Inference efficiency of RELISH}]{\new{\textbf{Inference efficiency of {\method}.} We report throughput as the number of examples processed per second, averaged across four LLM backbones and three random seeds. Latency/{\method} compares each baseline's wall-clock latency against {\method}, where larger values indicate slower inference. {\method} adds negligible inference overhead compared to simple predictive heads and is faster than autoregressive and regression-aware methods.}}
\label{tab:ablation_inference_time}
\end{table*}

\new{
Table~\ref{tab:ablation_inference_time} reports `Throughput (ex/s)` by dividing the number of corresponding test set examples by the total wall-clock time. Final metrics are then averaged across four LLMs and three seeds within each dataset to yield a method-level overview. For ease of comparison, we also show a `Latency/RELISH` column that compares each baseline’s relative throughput against RELISH.
}

\new{
We observe that RELISH adds negligible inference overhead compared to simple predictive heads (only $\sim1$\% slower). In contrast, autoregressive and regression-aware methods incur more overhead. On STS-B, these methods increase latency by factors ranging from 4.85x to 14.95x relative to RELISH. This gap widens on the WMT\_EN\_ZH dataset, where these baselines incur latency penalties ranging from 6.45x to 24.86x. Ultimately, these results suggest that RELISH is both efficient during training (fewer trainable parameters; see $\S$~\ref{sec:results_footprint}) and inference (high throughput). 
}

\subsection{\new{Does warm-starting $r^{(0)}$ improve {\method}?}}
\label{sec:app_ablations_warm_start}

\new{
For results reported in $\S$~\ref{sec:results}, we randomly initialized the learned latent state $r^{(0)}$ for all experiments. Another variant is to initialize $r^{(0)}$ from an input-dependent representation, which might give the refinement blocks a warm start and reduce the number of optimization steps needed.}

\new{
Toward this end, we compare {\method} with different warm-start configurations. 
Namely, we compare random (reported in $\S$~\ref{sec:results}) to last-token and mean-pooled initializations for $r^{(0)}$ on the STS-B and WMT\_EN\_ZH datasets across the same four LLMs and three seeds. 
We follow the same experimental setup details in $\S$~\ref{sec:experimental_setup_extended} and only change the initialization scheme. 
To measure whether warm-start can lead to faster convergence, we also report `Training Steps` as the number of `optim.step()` counts. 
Since we do not use gradient accumulation, one step roughly corresponds to one backpropagation step for each training batch. 
We use a batch size of 32 for all variants, so training steps are comparable across models within a dataset, but not directly across datasets since WMT\_EN\_ZH has more training data (7k vs. 5.7k).
For readability, we aggregate results over all LLMs and seeds for each configuration.
}

\begin{table*}[t]
\centering
\small
\setlength{\tabcolsep}{6pt}
\renewcommand{\arraystretch}{1.15}
\resizebox{\textwidth}{!}{%
\begin{tabular}{llcccc}
\toprule
\textbf{Dataset} & \textbf{{\method} Variant} & \textbf{RMSE} $\downarrow$ & \textbf{Pearson} $\uparrow$ & \textbf{Spearman} $\uparrow$ & \textbf{Training Steps} $\downarrow$ \\
\midrule
\multirow{3}{*}{STS-B} & Last Token & $0.677_{\scriptscriptstyle \pm 0.059}$ & $0.898_{\scriptscriptstyle \pm 0.019}$ & $0.897_{\scriptscriptstyle \pm 0.017}$ & $1035_{\scriptscriptstyle \pm 267}$ \\
                       & Mean Pool & $0.688_{\scriptscriptstyle \pm 0.051}$ & $0.894_{\scriptscriptstyle \pm 0.017}$ & $0.893_{\scriptscriptstyle \pm 0.016}$ & $\mathbf{960_{\scriptscriptstyle \pm 310}}$ \\
                       & Random (Ours) & $\mathbf{0.623_{\scriptscriptstyle \pm 0.037}}$ & $\mathbf{0.917_{\scriptscriptstyle \pm 0.010}}$ & $\mathbf{0.916_{\scriptscriptstyle \pm 0.009}}$ & $1080_{\scriptscriptstyle \pm 188}$ \\
\cmidrule(lr){1-6}
\multirow{3}{*}{WMT\_EN\_ZH} & Last Token & $\mathbf{11.088_{\scriptscriptstyle \pm 1.092}}$ & $0.501_{\scriptscriptstyle \pm 0.029}$ & $0.493_{\scriptscriptstyle \pm 0.025}$ & $493_{\scriptscriptstyle \pm 297}$ \\
                           & Mean Pool & $11.365_{\scriptscriptstyle \pm 1.206}$ & $0.498_{\scriptscriptstyle \pm 0.027}$ & $0.489_{\scriptscriptstyle \pm 0.024}$ & $493_{\scriptscriptstyle \pm 249}$ \\
                           & Random (Ours) & $11.410_{\scriptscriptstyle \pm 1.521}$ & $\mathbf{0.524_{\scriptscriptstyle \pm 0.040}}$ & $\mathbf{0.521_{\scriptscriptstyle \pm 0.030}}$ & $\mathbf{420_{\scriptscriptstyle \pm 174}}$ \\
\bottomrule
\end{tabular}
}
\caption{\new{\textbf{Warm start ablations for $r^{(0)}$.} We compare the default random initialization of $r^{(0)}$ against input-dependent last-token and mean-pooling warm start configurations. Metrics are averaged across four LLM backbones and three random seeds. Bold indicates the best variant within each dataset and metric. Overall, random initialization appears to be the most robust choice, matching or exceeding the predictive performance of other warm-start variants while maintaining training efficiency.}}
\label{tab:ablation_warm_start}
\end{table*}

\new{
Table~\ref{tab:ablation_warm_start} shows that input-dependent warm starts do not yield consistent benefits. Mean-pool initialization slightly reduces training steps on STS-B only, and last-token initialization slightly lowers RMSE on WMT\_EN\_ZH only. On the other hand, the default random initialization achieves the best Pearson and Spearman correlations on both datasets, the best RMSE on STS-B, and the fewest training steps on WMT\_EN\_ZH. We therefore retain the simpler random initialization as the most robust default.
}

\section{Limitations and Future Work}
\label{sec:app_limitations}

\paragraph{Reliance on LLM backbones.} 
Because {\method} operates over a frozen LLM backbone, its predictive capability is inherently bounded by the quality of the LLM's pre-trained representations. That is, {\method} is designed to robustly extract and refine regression-relevant signals that are \textit{already embedded} in model hidden states. Thus, for tasks that require novel or out-of-distribution domain knowledge, we would expect some fine-tuning of the LLM backbone to be necessary. 

\paragraph{Computational Scaling with Sequence Length.} 
{\method} adds minimal trainable parameters ($\sim$3.4--3.7M) and avoids the heavy inference costs of autoregressive generation or posterior approximation of regression-aware inference methods. However, its attention-based pooling mechanism scales linearly with the input sequence length. For exceptionally long inputs (e.g., estimating the quality of entire research papers), cross-attention operations over token-level representations could become a computational bottleneck. 

\paragraph{Scope of Evaluated Tasks.}
\new{Our evaluation covers semantic similarity, reference-free translation quality estimation, and code quality estimation. Future work may test how well {\method} generalizes to further regression tasks, such as continuous sentiment ratings, product prices, or average product ratings. Recent work by \citet{Vedula2025-yn} suggests product-related prediction as a promising direction, and its authors helpfully suggested two public resources---Amazon Reviews 2018\footnote{\url{https://cseweb.ucsd.edu/~jmcauley/datasets/amazon_v2/}} and 2023\footnote{\url{https://amazon-reviews-2023.github.io/}}---that could support such extensions.}



\paragraph{\new{Combining {\method} with RAFT.}}
\label{sec:app_relish_raft_hybrid}

\new{Since RAFT \citep{Lukasik2025-ak} performs better under sparse data settings, combining {\method} with RAFT is an interesting direction for future work. A simple extension may replace {\method}'s linear regressor with a RAFT-style decision-theoretic inference rule. Rather than linearly projecting the final latent state $r^{(L)}$ to a single scalar, the predictive head could project $r^{(L)}$ into logits over a grid of candidates, convert those logits into weights, and compute the final scalar as an expectation over grid values. This hybrid would preserve {\method}'s learned token-summary mechanism while using RAFT's grid-based predictor, but it would also reintroduce training and inference overhead that scales with the number of grid candidates.}

\paragraph{\new{Leveraging Chain of Thought Rationales.}}
\label{sec:app_cot_relish}

\new{While predictive head methods do not leverage CoT in the same way that autoregressive methods do, {\method} could still leverage CoT. One way to do this would be to incorporate CoT rationales into task inputs. Since {\method} operates on token representations, any rationales included in the prompt can be attended to by the learned latent query. Alternatively, instead of taking CoT rationales as inputs, one could generate them as outputs. For example, future work could explore a multi-task approach in which the LLM backbone is fine-tuned to output CoT rationales, while {\method} makes predictions conditioned on both the original input and generated traces.}

\new{Beyond the work required above to leverage CoT, evaluation would also need to be significantly expanded. A fair study would evaluate all regression baselines, with and without CoT rationales, across various datasets and LLMs. Beyond making relative comparisons between methods, one would need to carefully analyze how changes in performance stem from the CoT rationales rather than from the regression methods themselves. One would also need to decide which types of CoT rationales to evaluate: human annotation, self-generated by prompting the LLM backbone, or generated by another teacher LLM.}

\new{Finally, leveraging CoT could actually hurt performance. Prior work suggests that incorporating rationales requires careful design \citep{huang2023large}. \citet{Chiang2025-ax} show that fine-tuning on self-generated lower-quality rationales reduces performance. Similarly, systems like Quiet-STaR rely on filtering or rewarding useful rationales rather than indiscriminately training against all generated rationales \citep{zelikman2024quiet}. Our present study fairly compares all methods without CoT and, for the reasons above, leaves a robust investigation of CoT in regression to future work.}

\begin{table}
\centering
\small
\begin{tabular}{@{}lp{10.8cm}@{}}
\toprule
\rowcolor[HTML]{EFEFEF} \multicolumn{2}{l}{\textbf{Dataset: STS-B}} \\ \midrule
\textbf{System Prompt} & 
\begin{minipage}[t]{10.8cm}
\begin{spverbatim}
You are an expert human annotator for semantic textual similarity (STS). You will see English sentence pairs and judge how similar in MEANING they are.

Scoring instructions:
- Output a real-valued similarity score from 0.000 to 5.000, inclusive.
- 5.000 = same meaning / paraphrases.
- 2.500 = partially similar with important differences.
- 0.000 = completely unrelated meaning.

CRITICAL FORMAT RULE:
Return ONLY the numeric score, with EXACTLY three digits after the decimal point (e.g., 3.000). Do NOT include any words, punctuation, units, labels, or explanation.
\end{spverbatim} 
\end{minipage} \\ \addlinespace
\textbf{User Message} & 
\begin{minipage}[t]{10.8cm}
\begin{spverbatim}
Sentence A: {sentence_a}
Sentence B: {sentence_b}

Score (0.000 to 5.000 only):
\end{spverbatim}
\end{minipage} \\
\bottomrule
\end{tabular}
\caption{Full prompt template for the STS-Benchmark dataset.}
\label{tab:prompt_stsb}
\end{table}

\begin{table}
\centering
\small
\begin{tabular}{@{}lp{10.8cm}@{}}
\toprule
\rowcolor[HTML]{EFEFEF} \multicolumn{2}{l}{\textbf{Dataset: SICKR-STS}} \\ \midrule
\textbf{System Prompt} & 
\begin{minipage}[t]{10.8cm}
\begin{spverbatim}
You are an expert human annotator for semantic textual similarity (STS) for Sentences Involving Compositional Knowledge. You will see sentence pairs and judge how similar in MEANING they are.

Scoring instructions:
- Output a real-valued similarity score from 1.000 to 5.000, inclusive.
- 5.000 = same meaning / paraphrases; 1.000 = completely unrelated.

CRITICAL FORMAT RULE:
Return ONLY the numeric score, with EXACTLY three digits after the decimal point (e.g., 3.000). Do NOT include any words, punctuation, units, labels, or explanation.
\end{spverbatim}
\end{minipage} \\ \addlinespace
\textbf{User Message} & 
\begin{minipage}[t]{10.8cm}
\begin{spverbatim}
Sentence A: {sentence_a}
Sentence B: {sentence_b}

Score (1.000 to 5.000 only):
\end{spverbatim}
\end{minipage} \\
\bottomrule
\end{tabular}
\caption{Full prompt template for the SICKR-STS dataset.}
\label{tab:prompt_sickr}
\end{table}

\begin{table}
\centering
\small
\begin{tabular}{@{}lp{10.8cm}@{}}
\toprule
\rowcolor[HTML]{EFEFEF} \multicolumn{2}{l}{\textbf{Dataset: WMT\_EN\_ZH}} \\ \midrule
\textbf{System Prompt} & 
\begin{minipage}[t]{10.8cm}
\begin{spverbatim}
You are an expert human annotator for translation quality / semantic textual similarity (STS). You will see sentence pairs where Sentence A is in English (source) and Sentence B is in Chinese (translation). You must judge how well the meaning is preserved from English to Chinese.

Scoring instructions:
- Output a real-valued translation quality / similarity score from 0.000 to 100.000.
- 100.000 = perfect meaning preservation (excellent translation / paraphrases).
- 50.000  = partially similar meaning but with important differences or missing info.
- 0.000   = completely unrelated or very bad translation.

CRITICAL FORMAT RULE:
Return ONLY the numeric score with EXACTLY three digits after the decimal point. Do NOT include any words, punctuation, units, labels, or explanation.
\end{spverbatim}
\end{minipage} \\ \addlinespace
\textbf{User Message} & 
\begin{minipage}[t]{10.8cm}
\begin{spverbatim}
Sentence A (English source): {sentence_a}
Sentence B (Chinese translation): {sentence_b}

Score (0.000 to 100.000 only):
\end{spverbatim}
\end{minipage} \\
\bottomrule
\end{tabular}
\caption{Full prompt template for the WMT English-Chinese quality estimation task.}
\label{tab:prompt_wmt_enzh}
\end{table}

\begin{table}
\centering
\small
\begin{tabular}{@{}lp{10.8cm}@{}}
\toprule
\rowcolor[HTML]{EFEFEF} \multicolumn{2}{l}{\textbf{Dataset: WMT\_RU\_EN}} \\ \midrule
\textbf{System Prompt} & 
\begin{minipage}[t]{10.8cm}
\begin{spverbatim}
You are an expert human annotator for translation quality / semantic textual similarity (STS). You will see sentence pairs where Sentence A is in Russian (source) and Sentence B is in English (translation). You must judge how well the meaning is preserved from Russian to English.

Scoring instructions:
- Output a real-valued translation quality / similarity score from 0.000 to 100.000.
- 100.000 = perfect meaning preservation (excellent translation / paraphrases).
- 50.000  = partially similar meaning but with important differences or missing info.
- 0.000   = completely unrelated or very bad translation.

CRITICAL FORMAT RULE:
Return ONLY the numeric score with EXACTLY three digits after the decimal point. Do NOT include any words, punctuation, units, labels, or explanation.
\end{spverbatim}
\end{minipage} \\ \addlinespace
\textbf{User Message} & 
\begin{minipage}[t]{10.8cm}
\begin{spverbatim}
Sentence A (Russian source): {sentence_a}
Sentence B (English translation): {sentence_b}

Score (0.000 to 100.000 only):
\end{spverbatim}
\end{minipage} \\
\bottomrule
\end{tabular}
\caption{Full prompt template for the WMT Russian-English quality estimation task.}
\label{tab:prompt_wmt_ruen}
\end{table}

\begin{table}
\centering
\small
\begin{tabular}{@{}lp{10.8cm}@{}}
\toprule
\rowcolor[HTML]{EFEFEF} \multicolumn{2}{l}{\textbf{Dataset: WMT\_SI\_EN}} \\ \midrule
\textbf{System Prompt} & 
\begin{minipage}[t]{10.8cm}
\begin{spverbatim}
You are an expert human annotator for translation quality / semantic textual similarity (STS). You will see sentence pairs where Sentence A is in Sinhala (source) and Sentence B is in English (translation). You must judge how well the meaning is preserved from Sinhala to English.

Scoring instructions:
- Output a real-valued translation quality / similarity score from 0.000 to 100.000.
- 100.000 = perfect meaning preservation (excellent translation / paraphrases).
- 50.000  = partially similar meaning but with important differences or missing info.
- 0.000   = completely unrelated or very bad translation.

CRITICAL FORMAT RULE:
Return ONLY the numeric score with EXACTLY three digits after the decimal point. Do NOT include any words, punctuation, units, labels, or explanation.
\end{spverbatim}
\end{minipage} \\ \addlinespace
\textbf{User Message} & 
\begin{minipage}[t]{10.8cm}
\begin{spverbatim}
Sentence A (Sinhala source): {sentence_a}
Sentence B (English translation): {sentence_b}

Score (0.000 to 100.000 only):
\end{spverbatim}
\end{minipage} \\
\bottomrule
\end{tabular}
\caption{Full prompt template for the WMT Sinhala-English quality estimation task.}
\label{tab:prompt_wmt_sien}
\end{table}

\begin{table}
\centering
\small
\begin{tabular}{@{}lp{10.8cm}@{}}
\toprule
\rowcolor[HTML]{EFEFEF} \multicolumn{2}{l}{\textbf{Dataset: StatLLM}} \\ \midrule
\textbf{System Prompt} & 
\begin{minipage}[t]{10.8cm}
\begin{spverbatim}
You are a human expert in statistical programming using the Statistical Analysis System (SAS) language. You will see a statistical analysis task, dataset description, and SAS code. You must judge the quality of the provided SAS code within its intended context.

Scoring instructions:
- Output a real-valued total score from 0.000 to 50.000, inclusive.
- 50.000 = excellent code: correct, executable, readable, concise, and produces the necessary clear output.
- 25.000 = partially correct code: addresses some of the task but has important mistakes, omissions, or unclear output.
- 0.000  = unusable code: does not address the task, is likely non-executable, or would produce wrong/missing output.

CRITICAL FORMAT RULE:
Return ONLY the numeric score with EXACTLY three digits after the decimal point. Do NOT include any words, punctuation, units, labels, or explanation.
\end{spverbatim}
\end{minipage} \\ \addlinespace
\textbf{User Message} & 
\begin{minipage}[t]{10.8cm}
\begin{spverbatim}
Statistical analysis task:
{problem_description}

Dataset name: {dataset_name}

Dataset description:
{data_description}

Generated SAS code:
```sas
{generated_sas_code}
```

Total score (0.000 to 50.000 only):
\end{spverbatim}
\end{minipage} \\
\bottomrule
\end{tabular}
\caption{\new{Full prompt template for the StatLLM SAS code quality estimation task.}}
\label{tab:prompt_statllm}
\end{table}

\begin{table*}[t]
\centering
\caption{Comparison on \textbf{STS-B} (Semantic Textual Similarity Benchmark). Metrics are mean $\pm$ std over 3 seeds and rounded to 3 decimals. Lower is better for RMSE ($\downarrow$); higher is better for correlations ($\uparrow$).}
\label{tab:stsb_relish_main}
\small
\setlength{\tabcolsep}{6pt}
\renewcommand{\arraystretch}{1.15}

\begin{tabularx}{\textwidth}{@{} l X c c c @{}}
\toprule
\multicolumn{5}{c}{\textbf{Dataset:} \textbf{STS-Benchmark} (Semantic Textual Similarity Benchmark)} \\
\midrule
\textbf{LLM encoder} & \textbf{Inference} & {\textbf{RMSE} $\downarrow$} & {\textbf{Pearson} $\uparrow$} & {\textbf{Spearman} $\uparrow$} \\
\midrule

\multirow{11}{*}{\textsc{Llama}~3.1~8B~Instruct} & Zero Shot & $1.215_{\scriptscriptstyle \pm 0.010}$ & $0.717_{\scriptscriptstyle \pm 0.006}$ & $0.688_{\scriptscriptstyle \pm 0.010}$ \\
& Many Shot & $1.235_{\scriptscriptstyle \pm 0.045}$ & $0.675_{\scriptscriptstyle \pm 0.031}$ & $0.635_{\scriptscriptstyle \pm 0.030}$ \\
& Linear & $1.323_{\scriptscriptstyle \pm 0.000}$ & $0.632_{\scriptscriptstyle \pm 0.000}$ & $0.637_{\scriptscriptstyle \pm 0.000}$ \\
& MLP & $1.051_{\scriptscriptstyle \pm 0.015}$ & $0.732_{\scriptscriptstyle \pm 0.009}$ & $0.725_{\scriptscriptstyle \pm 0.010}$ \\
& RAIL & $1.252_{\scriptscriptstyle \pm 0.007}$ & $0.743_{\scriptscriptstyle \pm 0.002}$ & $0.771_{\scriptscriptstyle \pm 0.003}$ \\
\cdashline{2-5}[0.4pt/2pt]
& LoRA + Zero Shot & $0.788_{\scriptscriptstyle \pm 0.032}$ & $0.858_{\scriptscriptstyle \pm 0.014}$ & $0.853_{\scriptscriptstyle \pm 0.017}$ \\
\bottomrule
\end{tabularx}
\end{table*}

\begin{table*}[t]
\centering
\caption{Comparison on \textbf{SICKR\_STS} (Sentences Involving Compositional Knowledge, Revised). Metrics are mean $\pm$ std over 3 seeds and rounded to 3 decimals. Lower is better for RMSE ($\downarrow$); higher is better for correlations ($\uparrow$).}
\label{tab:sickr_sts_relish_main}
\small
\setlength{\tabcolsep}{6pt}
\renewcommand{\arraystretch}{1.15}

\begin{tabularx}{\textwidth}{@{} l X c c c @{}}
\toprule
\multicolumn{5}{c}{\textbf{Dataset:} \textbf{SICKR\_STS} (Sentences Involving Compositional Knowledge, Revised)} \\
\midrule
\textbf{LLM encoder} & \textbf{Inference} & {\textbf{RMSE} $\downarrow$} & {\textbf{Pearson} $\uparrow$} & {\textbf{Spearman} $\uparrow$} \\
\midrule

\multirow{11}{*}{\textsc{Llama}~3.1~8B~Instruct} & Zero Shot & $1.095_{\scriptscriptstyle \pm 0.012}$ & $0.666_{\scriptscriptstyle \pm 0.015}$ & $0.650_{\scriptscriptstyle \pm 0.014}$ \\
& Many Shot & $0.918_{\scriptscriptstyle \pm 0.090}$ & $0.608_{\scriptscriptstyle \pm 0.019}$ & $0.477_{\scriptscriptstyle \pm 0.036}$ \\
& Linear & $0.612_{\scriptscriptstyle \pm 0.000}$ & $0.807_{\scriptscriptstyle \pm 0.000}$ & $0.764_{\scriptscriptstyle \pm 0.000}$ \\
& MLP & $0.509_{\scriptscriptstyle \pm 0.007}$ & $0.858_{\scriptscriptstyle \pm 0.004}$ & $0.809_{\scriptscriptstyle \pm 0.007}$ \\
& RAIL & $1.126_{\scriptscriptstyle \pm 0.003}$ & $0.686_{\scriptscriptstyle \pm 0.002}$ & $0.746_{\scriptscriptstyle \pm 0.002}$ \\
\cdashline{2-5}[0.4pt/2pt]
& LoRA + Zero Shot & $0.489_{\scriptscriptstyle \pm 0.017}$ & $0.878_{\scriptscriptstyle \pm 0.008}$ & $0.822_{\scriptscriptstyle \pm 0.007}$ \\
\bottomrule
\end{tabularx}
\end{table*}

\begin{table*}[t]
\centering
\caption{Comparison on \textbf{WMT\_RU\_EN} (Russian $\rightarrow$ English). Metrics are mean $\pm$ std over 3 seeds and rounded to 3 decimals. Lower is better for RMSE ($\downarrow$); higher is better for correlations ($\uparrow$).}
\label{tab:wmt_en_ru_relish_main}
\small
\setlength{\tabcolsep}{6pt}
\renewcommand{\arraystretch}{1.15}

\begin{tabularx}{\textwidth}{@{} l X c c c @{}}
\toprule
\multicolumn{5}{c}{\textbf{Dataset:} \textbf{WMT\_RU\_EN} (Russian $\rightarrow$ English)} \\
\midrule
\textbf{LLM encoder} & \textbf{Inference} & {\textbf{RMSE} $\downarrow$} & {\textbf{Pearson} $\uparrow$} & {\textbf{Spearman} $\uparrow$} \\
\midrule

\multirow{11}{*}{\textsc{Llama}~3.1~8B~Instruct} & Zero Shot & $32.853_{\scriptscriptstyle \pm 0.160}$ & $0.473_{\scriptscriptstyle \pm 0.008}$ & $0.409_{\scriptscriptstyle \pm 0.012}$ \\
& Many Shot & $31.481_{\scriptscriptstyle \pm 1.342}$ & $0.198_{\scriptscriptstyle \pm 0.106}$ & $0.133_{\scriptscriptstyle \pm 0.098}$ \\
& Linear & $22.637_{\scriptscriptstyle \pm 0.000}$ & $0.619_{\scriptscriptstyle \pm 0.000}$ & $0.567_{\scriptscriptstyle \pm 0.000}$ \\
& MLP & $19.161_{\scriptscriptstyle \pm 0.045}$ & $0.695_{\scriptscriptstyle \pm 0.001}$ & $0.646_{\scriptscriptstyle \pm 0.004}$ \\
& RAIL & $33.237_{\scriptscriptstyle \pm 0.175}$ & $0.528_{\scriptscriptstyle \pm 0.005}$ & $0.530_{\scriptscriptstyle \pm 0.007}$ \\
\cdashline{2-5}[0.4pt/2pt]
& LoRA + Zero Shot & $20.264_{\scriptscriptstyle \pm 0.426}$ & $0.718_{\scriptscriptstyle \pm 0.018}$ & $0.633_{\scriptscriptstyle \pm 0.018}$ \\
\bottomrule
\end{tabularx}
\end{table*}

\begin{table*}[t]
\centering
\caption{Comparison on \textbf{WMT\_EN\_ZH} (English $\rightarrow$ Chinese). Metrics are mean $\pm$ std over 3 seeds and rounded to 3 decimals. Lower is better for RMSE ($\downarrow$); higher is better for correlations ($\uparrow$).}
\label{tab:wmt_en_zh_relish_main}
\small
\setlength{\tabcolsep}{6pt}
\renewcommand{\arraystretch}{1.15}

\begin{tabularx}{\textwidth}{@{} l X c c c @{}}
\toprule
\multicolumn{5}{c}{\textbf{Dataset:} \textbf{WMT\_EN\_ZH} (English $\rightarrow$ Chinese)} \\
\midrule
\textbf{LLM encoder} & \textbf{Inference} & {\textbf{RMSE} $\downarrow$} & {\textbf{Pearson} $\uparrow$} & {\textbf{Spearman} $\uparrow$} \\
\midrule

\multirow{11}{*}{\textsc{Llama}~3.1~8B~Instruct} & Zero Shot & $33.543_{\scriptscriptstyle \pm 0.217}$ & $0.348_{\scriptscriptstyle \pm 0.013}$ & $0.276_{\scriptscriptstyle \pm 0.018}$ \\
& Many Shot & $26.312_{\scriptscriptstyle \pm 1.674}$ & $0.075_{\scriptscriptstyle \pm 0.010}$ & $0.042_{\scriptscriptstyle \pm 0.013}$ \\
& Linear & $19.648_{\scriptscriptstyle \pm 0.000}$ & $0.277_{\scriptscriptstyle \pm 0.000}$ & $0.253_{\scriptscriptstyle \pm 0.000}$ \\
& MLP & $14.198_{\scriptscriptstyle \pm 0.057}$ & $0.331_{\scriptscriptstyle \pm 0.012}$ & $0.321_{\scriptscriptstyle \pm 0.011}$ \\
& RAIL & $38.840_{\scriptscriptstyle \pm 0.096}$ & $0.345_{\scriptscriptstyle \pm 0.004}$ & $0.367_{\scriptscriptstyle \pm 0.002}$ \\
\cdashline{2-5}[0.4pt/2pt]
& LoRA + Zero Shot & $17.827_{\scriptscriptstyle \pm 0.338}$ & $0.345_{\scriptscriptstyle \pm 0.033}$ & $0.335_{\scriptscriptstyle \pm 0.003}$ \\
\bottomrule
\end{tabularx}
\end{table*}

\begin{table*}[t]
\centering
\caption{Comparison on \textbf{WMT\_SI\_EN} (Sinhala $\rightarrow$ English). Metrics are mean $\pm$ std over 3 seeds and rounded to 3 decimals. Lower is better for RMSE ($\downarrow$); higher is better for correlations ($\uparrow$).}
\label{tab:wmt_si_en_relish_main}
\small
\setlength{\tabcolsep}{6pt}
\renewcommand{\arraystretch}{1.15}

\begin{tabularx}{\textwidth}{@{} l X c c c @{}}
\toprule
\multicolumn{5}{c}{\textbf{Dataset:} \textbf{WMT\_SI\_EN} (Sinhala $\rightarrow$ English)} \\
\midrule
\textbf{LLM encoder} & \textbf{Inference} & {\textbf{RMSE} $\downarrow$} & {\textbf{Pearson} $\uparrow$} & {\textbf{Spearman} $\uparrow$} \\
\midrule

\multirow{11}{*}{\textsc{Llama}~3.1~8B~Instruct} & Zero Shot & $40.343_{\scriptscriptstyle \pm 0.720}$ & $0.433_{\scriptscriptstyle \pm 0.028}$ & $0.427_{\scriptscriptstyle \pm 0.020}$ \\
& Many Shot & $35.311_{\scriptscriptstyle \pm 3.971}$ & $0.186_{\scriptscriptstyle \pm 0.020}$ & $0.149_{\scriptscriptstyle \pm 0.054}$ \\
& Linear & $29.768_{\scriptscriptstyle \pm 0.000}$ & $0.368_{\scriptscriptstyle \pm 0.000}$ & $0.348_{\scriptscriptstyle \pm 0.000}$ \\
& MLP & $19.460_{\scriptscriptstyle \pm 0.198}$ & $0.521_{\scriptscriptstyle \pm 0.004}$ & $0.477_{\scriptscriptstyle \pm 0.004}$ \\
& RAIL & $44.966_{\scriptscriptstyle \pm 0.055}$ & $0.438_{\scriptscriptstyle \pm 0.001}$ & $0.481_{\scriptscriptstyle \pm 0.010}$ \\
\cdashline{2-5}[0.4pt/2pt]
& LoRA + Zero Shot & $28.894_{\scriptscriptstyle \pm 1.483}$ & $0.448_{\scriptscriptstyle \pm 0.041}$ & $0.438_{\scriptscriptstyle \pm 0.046}$ \\
\bottomrule
\end{tabularx}
\end{table*}

\begin{table*}[t]
\centering
\caption{\new{Comparison on \textbf{StatLLM} (code quality estimation). Metrics are mean $\pm$ std over 3 seeds and rounded to 3 decimals. Lower is better for RMSE ($\downarrow$); higher is better for correlations ($\uparrow$).}}
\label{tab:statllm_code_quality_relish_main}
\small
\setlength{\tabcolsep}{6pt}
\renewcommand{\arraystretch}{1.15}

\begin{tabularx}{\textwidth}{@{} l X c c c @{}}
\toprule
\multicolumn{5}{c}{\textbf{Dataset:} \textbf{StatLLM}} \\
\midrule
\textbf{LLM encoder} & \textbf{Inference} & {\textbf{RMSE} $\downarrow$} & {\textbf{Pearson} $\uparrow$} & {\textbf{Spearman} $\uparrow$} \\
\midrule

\multirow{11}{*}{\textsc{Llama}~3.1~8B~Instruct} & Zero Shot & $26.170_{\scriptscriptstyle \pm 0.121}$ & $0.060_{\scriptscriptstyle \pm 0.025}$ & $0.038_{\scriptscriptstyle \pm 0.013}$ \\
& Many Shot & $13.988_{\scriptscriptstyle \pm 0.777}$ & $0.052_{\scriptscriptstyle \pm 0.072}$ & $0.067_{\scriptscriptstyle \pm 0.078}$ \\
& Linear & $20.478_{\scriptscriptstyle \pm 0.000}$ & $0.022_{\scriptscriptstyle \pm 0.000}$ & $0.032_{\scriptscriptstyle \pm 0.000}$ \\
& MLP & $23.720_{\scriptscriptstyle \pm 4.167}$ & $0.158_{\scriptscriptstyle \pm 0.000}$ & $0.153_{\scriptscriptstyle \pm 0.000}$ \\
& RAIL & $27.078_{\scriptscriptstyle \pm 0.112}$ & $0.020_{\scriptscriptstyle \pm 0.021}$ & $0.039_{\scriptscriptstyle \pm 0.019}$ \\
\cdashline{2-5}[0.4pt/2pt]
& LoRA + Zero Shot & $15.838_{\scriptscriptstyle \pm 1.675}$ & $0.222_{\scriptscriptstyle \pm 0.063}$ & $0.176_{\scriptscriptstyle \pm 0.088}$ \\
\bottomrule
\end{tabularx}
\end{table*}

\end{document}